\pdfoutput=1

\documentclass[11pt]{article}
\usepackage{CJKutf8}
\usepackage{acl}

\usepackage{times}
\usepackage{latexsym}
\usepackage{multirow}
\usepackage{multicol}
\usepackage{amsmath}
\usepackage{booktabs}
\usepackage{xcolor}
\usepackage{float}
\usepackage{algpseudocode}
\usepackage{listings}
\usepackage{color}
\usepackage{footnote}
\DeclareCaptionLabelFormat{andtable}{#1~#2  \&  \tablename~\thetable}
\usepackage{graphicx}
\usepackage[flushleft]{threeparttable}
\definecolor{backcolour}{rgb}{0.95, 0.95, 0.96}
\lstdefinestyle{mystyle}{
    backgroundcolor=\color{backcolour},   
    showtabs=false,                  
    tabsize=2
}

\usepackage{enumitem}

\usepackage[T1]{fontenc}

\usepackage[utf8]{inputenc}

\usepackage[scaled=0.8]{beramono}

\usepackage{microtype}

\usepackage[normalem]{ulem}

\newcommand{\rd}[1]{\textcolor{red}{#1}}
\newcommand{\bc}[1]{\textcolor{blue}{#1}}
\newcommand\cncpt{\textit{Country/Concept}}
\newcommand\pref{\textit{Country/Prefix}}
\newcommand\expt{\textit{Expert Units}}
\newcommand\georep{Geographic-Representation}
\newcommand\neigh{Neighbourhood Score}
\newcommand\reps{Representation Score}

\usepackage{tikz}
\usepackage{tikzscale}
\usepackage{pgfplots}
\usetikzlibrary{trees}

\title{Geographic and Geopolitical Biases of Language Models}

\author{Fahim Faisal, Antonios Anastasopoulos\\
Department of Computer Science, George Mason University\\
\texttt{\{ffaisal,antonis\}@gmu.edu}}

\begin{document}
\maketitle
\begin{abstract}
Pretrained language models (PLMs) often fail to fairly represent target users from certain world regions because of the under-representation of those regions in training datasets.  With recent PLMs trained on enormous data sources, quantifying their potential biases is difficult, due to their black-box nature and the sheer scale of the data sources. In this work, we devise an approach to study the geographic bias (and knowledge) present in PLMs, proposing a Geographic-Representation Probing Framework adopting a self-conditioning method coupled with entity-country mappings. Our findings suggest PLMs' representations map surprisingly well to the physical world in terms of country-to-country associations, but this knowledge is unequally shared across languages. Last, we explain how large PLMs despite exhibiting notions of geographical proximity, over-amplify geopolitical favouritism at inference time.\footnote{Code and data to reproduce our study will be made publicly available.}
\end{abstract}

\section{Introduction}

Large pretrained language models (PLMs) are capable of generating meaningful texts beyond English and very likely, models like GPT-3~\cite{brown2020language, https://doi.org/10.48550/arxiv.2204.07580, zhang2022opt, bloom} will form the go-to base model for automating tasks like summarizing texts, generating datasets given certain instructions \cite{schick-schutze-2021-generating} or perhaps even evaluating the generated texts~\cite{NEURIPS2021_e4d2b6e6}. While these PLMs continue to expand their utility, it is crucial that one also examines the potential biases that these PLMs exhibit. Moreover, the utility of these PLMs should be equitable to their target users so that they perform evenly for all speakers of the languages it is primarily trained on. Otherwise, the disparity that lies in the model (if any) will propagate further. To better illustrate these dynamics, consider a $L_1$ Spanish speaker from Peru, who is using a prompt-based PLM (like that of~\citet{wang-etal-2022-promda, https://doi.org/10.48550/arxiv.2109.09193}) to generate a localized synthetic dataset for some downstream task. They may use Spanish \textit{as used in the local context} to form their seed data/prefix/prompts. Now, if this language model has already skewed preferences towards geopolitically important countries, it is likely the generated texts will reflect this skewness, thus not appropriately reflecting the local, Peruvian context that the practitioner is interested in. 
\begin{figure}[t]
    \centering
    \begin{tabular}{c}
       \hspace{-4em} \includegraphics[width=.5\textwidth]{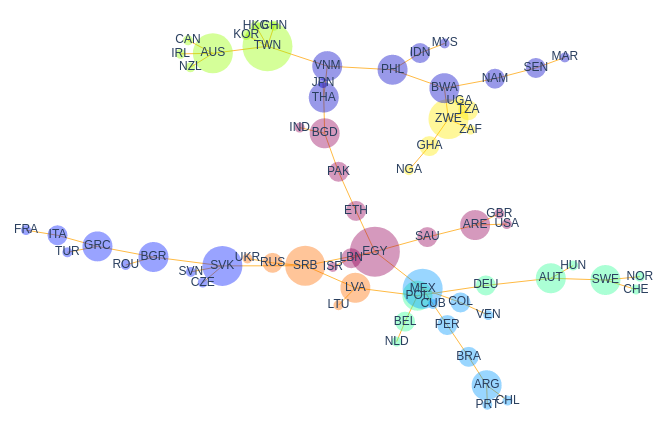} \vspace{-3em}\\
        \includegraphics[width=.4\textwidth]{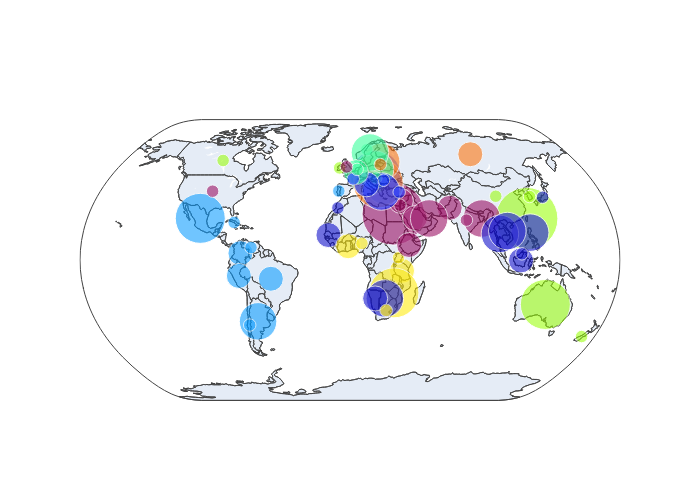}
    \end{tabular}
    \vspace{-3em}
    \caption{Example of a Geographic Representation network and it's corresponding location clusters (colored) recovered from the top-$50$ country-"expert" neurons of \texttt{BLOOM}. Notice that connected countries are either geographically or culturally close (e.g. south American cluster in light blue, African countries in yellow, South-East Asian countries in dark blue). \textit{Note: node size is proportional to its degree in the graph.}}
    \label{fig:main_ex}
    \vspace{-1em}
\end{figure}
However, the quantification of this presumed geographic disparity in PLMs is not yet explored. Though given the well-documented western-country bias (or Global North bias) exhibited in most NLP benchmarks and datasets~\cite[\textit{inter alia}]{faisal-etal-2022-dataset}, we hypothesize that text generation models might also suffer from the similar pitfall. On top of this, how language variety impact the distribution of geographic knowledge encoded in PLMs (eg. sense of country-country association in physical or geopolitical space) is also under-explored.

Herein, we perform an evidence-based study to unfold the underlying geographic distribution of multilingual PLMs. We propose a pipeline to probe the Text-Generative PLMs using prompt-based inference for Geographic-Knowledge as well as existing domain-variant disparity (geography in our case). Our research questions and key findings are:

\begin{itemize}[noitemsep,nolistsep,leftmargin=*]
  \item \textbf{RQ1:} \textit{To what extent is geographic proximity encoded in the PLMs?} \textbf{F:} PLMs can infer geographic proximity surprisingly well in terms of country-country association (see Figure \ref{fig:main_ex}). However, we also observe over-representation of certain countries during text generation.
    
     \item \textbf{RQ2:} \textit{What is the influence of multilinguality in PLM's knowledge distribution of geographic proximity?} \textbf{F:} The shared multilingual representation space of PLMs has an uneven distribution of knowledge across languages. 
     
     \item \textbf{RQ3:} \textit{What is the effect of prompting using a geographic identifier (eg. \textit{"In Colombia"} <generate text>) on multilingual text generation?} \textbf{F:} Prompting with certain geographic identifiers can even alter the language of free-form generated text.
\end{itemize}

\section{Background and Related Work}
\label{sec:background}
A substantial amount of work has investigated existing social bias (eg. gender, racial, ethnic, occupational) identification and mitigation approaches in PLMs including, reducing token sensitivity during text generation~\cite{https://doi.org/10.48550/arxiv.2106.13219}, investigating model sensitivity \cite{immer-etal-2022-probing}, prompting using natural sentences~\cite{alnegheimish-etal-2022-using} and probing via embedding lookup \cite{ahn-oh-2021-mitigating}. On the other hand, a number of studies experimented with the behavior different PLMs exhibits while probing with geographic-context as well as cultural-commonsense   \cite{yin2022geomlama, ghosh-etal-2021-detecting}. 
However, we need to extract the specific model weights responsible for these observable polarity. Then using those weights in a controlled setting, we might be able to unfold how PLMs encode geographic knowledge as well as explain the exhibition of geographic-bias during inference.


\paragraph{Self Conditioning Pretrained Models}
\citet{suau2022selfcond} propose one such approach that extracts PLM weights having certain polarity and then prioritize those weights during text generation. Based on the generated text, they can quantify gender and occupation bias encoded by the PLM. As an example, consider a binary sentence classification task where positive class examples contain the mention of a concept word (eg. doctor) and vice-versa. A PLM is able to provide scores to these positive and negative examples. Looking at the average precision scores and the scores given by different model weights from each layer, we can identify the ones providing higher scores towards the positive examples. \citet{suau2022selfcond} refer to these model weights as \textit{expert units}. 

With these expert units identified, they can be further prioritized during text generation just by setting their expected values as if the concept word "doctor" was present in the generated text. This allows the PLM to generate texts relevant to the concept word without it explicitly mentioned. In the work of \citet{suau2022selfcond}, by comparing the generated texts, they easily quantify the presence of gender-specific words thus evaluating the presence of gender bias in the PLM (for example, consider the number of sentences where the context relates to the word "doctor" and mentions male-gender words compared to female-gender words). This approach serves two main purposes: (1) Identifying expert units: model parameters responsible for generating text related to the target concept (i.e. doctor). (2) Triggering specific behaviour in text generation without explicit mentioning of the target context, which inadvertently influences the behaviour of the model.



\section{Geographic Representation Probing}
\begin{figure*}[t]
    \centering
    \includegraphics[width=.99\textwidth]{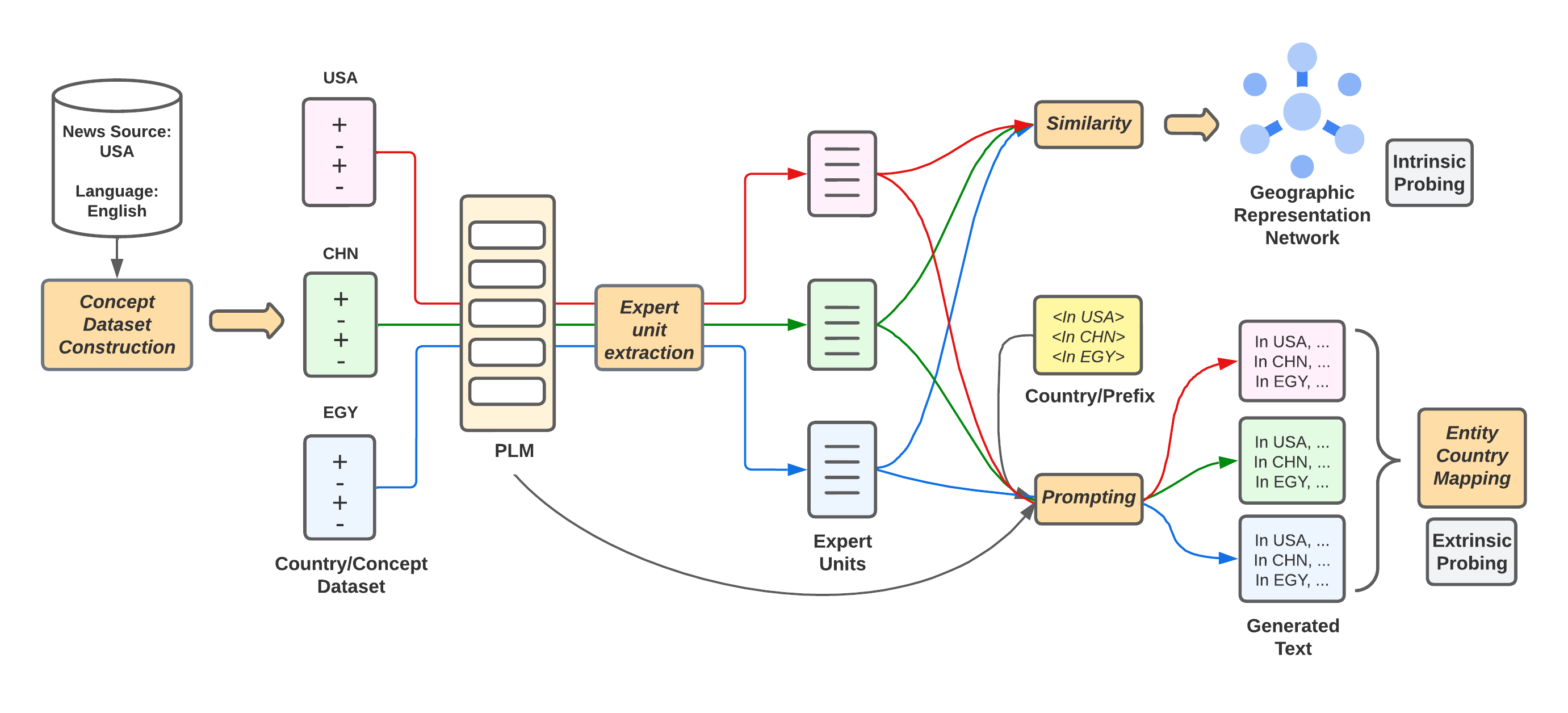}
    \vspace{-2em}
    \caption{Geographic Representation Probing Framework. We start with constructing the \textit{Country/Concept} dataset. Then we extract \textit{Expert Units} from the base PLM. After that, we use similarity measurement to prepare our Geographic Representation Network to perform Intrinsic Probing. In Parallel, we prompt the self-conditioned PLM with Geographic Identifiers (i.e. \textit{Country/Prefix}). Finally, we map the generated-text entities to countries to perform Extrinsic Probing.
}
    \label{tab:grn}
    \vspace{-1em}
\end{figure*}

In our study, we use the idea of Self Conditioning Pretrained Models to first extract expert units (i.e. model weights) which encode geographic knowledge and then we use those units during prompt-based text generation having different geographic identifier mentions. An example: Using some sentences with the mention as well as absence of the word \textit{"China"} to extract expert units and then, prioritize these units during text generation with the prompt \textit{"In USA ..."}. The aim here is to simulate an environment where we evaluate the model knowledge (\cncpt{}-specific \expt{}) by asking what it knows about other countries (i.e \pref{}). This allows us to quantify existing geographic bias and favouritism towards certain attributes present in the representation space of multilingual PLMs. 

Our probing framework contains five steps (see Figure \ref{tab:grn}): (1) Concept Dataset Construction (2) Expert Unit Extraction (3) \georep{} Network Construction (4) Prompt-based Text Generation (5) Entity Country Mapping.

\paragraph{Concept Dataset Construction}
First of all, we prepare our concept dataset in a binary classification fashion using which, we later perform self-conditioning a PLM on geographic concepts. To make it quantifiable, we define country to be our main unit of reference and construct concept datasets where each "concept" is loosely centered around a country. An additional requirement for these datasets is that the data have not been used as part of the pretraining data of the PLMs. Hence, we turn to recent news articles (scrapped using Google news api\footnote{\url{https://github.com/ranahaani/GNews}}): as we can control the date on which these data became public, we can be sure that they were not used in any pre-training process (so far). Such a dataset should also allow us to get a reasonable representation of current geopolitical affairs. Depending on the news-source country and language, we build several such \cncpt{} datasets.  A \cncpt{} dataset \bc{\texttt{\{C\}-\{l\}}} contains news about several (${c_1,c_2,..c_i..c_n}$) countries in \texttt{\{l\}} language where the news-source is \texttt{\{C\}} country. Here, each \cncpt{} $c_i$ has 100 positive examples (mention of $c_i$) sentences and 300 negative examples (no mention) sentences. For example, \texttt{USA-eng} \cncpt{} dataset contains data from US sources, in English, which either mention other countries (there are 100 positive examples for each country $c_i$) or are random sentences not mentioning any countries (negative examples).

\paragraph{Expert Unit Extraction}
Using the self-conditioning framework, we identify high performing \expt{} for each \cncpt{}. For example, Consider the \cncpt{} \texttt{India} from the dataset \texttt{USA-eng}. The \expt{} are the model weights which provide higher scores for the presence of a concept (i.e. positive examples mentioning "India"). Observing the average precision scores, we select the top-$k$ (eg. 10, 50) \expt{} from each PLM layer.




\paragraph{\georep{} Network} Now utilizing all these model \expt{}, we construct our \georep{} Networks. We use jaccard similarity to measure the similarity between any given \cncpt{} pairs $c_i$ and $c_j$ and their corresponding \expt{}. Then, utilizing these similarity measurement scores as edges in a graph (the countries being the nodes), we prepare a PLM-specific Geographic Representation network for each of our \expt{} set. This network is a Minimum-Spanning Tree graph highlighting the internal country-country associations. We further make it easier to digest by identifying the community clusters of countries using the Louvain Community Detection method \cite{Blondel_2008}. For example, in Figure~\ref{fig:main_ex}, we show the network obtained with the \texttt{USA-eng} dataset from the BLOOM~\cite{bloom} \expt{}. Effectively, we can recover a very good geographical representation of the countries straight from the weights of the network.



    

\begin{figure}[t]
    \centering
    \includegraphics[width=.48\textwidth]{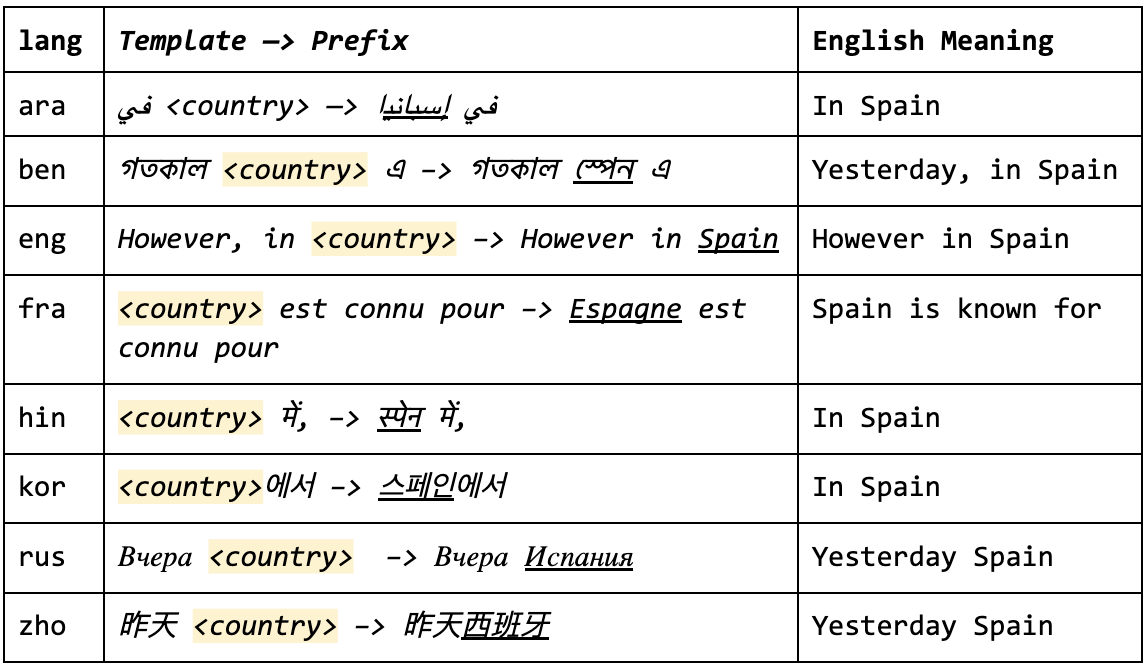}
    \vspace{-2em}
    \caption{Prefix construction using Multilingual Prefix-Templates. Here we replace the specific \textbf{\texttt{<country>}} position with \texttt{"Spain"} in the given language. See Appendix~\ref{app:prefix} for the complete list of multilingual prefix templates.
}
    \label{tab:template_main}
    \vspace{-1em}
\end{figure}

\paragraph{Prompt-based Text Generation}
With the \cncpt{}-specific \expt{} at hand, we can now investigate what happens when we use the PLM for text generation. The self-conditioning method~\cite{suau2022selfcond} uses sequential decoding and prioritize the \expt{} by approximating their scores from the average precision values predicted for a certain \cncpt{}. 
This allows us to artificially simulate the presence of a country name and it's related context during text generation. Now we perform text generation with one more twist: we provide one country-mention as part of the prefix/prompt (i.e. \pref{}). The idea here is to simulate an environment where we evaluate the model knowledge (\cncpt{}-specific \expt{}) by asking what it knows about other countries (i.e \pref{}). We generate several template-based multilingual prompts (the prefix construction process is depicted in Table \ref{tab:template_main}) where we replace the \texttt{<country>} tag with different country names. 

\paragraph{Entity Country Mapping}
Finally, to investigate the existence of geopolitical favouritism, we quantify the geographic biases of the generated texts by mapping any entities appearing in the text to corresponding countries. We use the Dataset Geography framework of~\citet{faisal-etal-2022-dataset}, which uses the multilingual entity linker mGENRE~\cite{de-cao-etal-2022-multilingual} for linking entity to Wikidata entries which are then mapped to countries.

\section{Experimental Settings}


    
    



\paragraph{Models and Languages}
We use GPT2-medium~\cite{radford2019language}, mGPT~\cite{https://doi.org/10.48550/arxiv.2204.07580} and BLOOM-560m~\cite{bloom}, all models available through huggingface. For the English dataset sourced from the US-News Platform \texttt{(USA-eng)} we extract \expt{} from all three models. For non-English datasets, we perform \expt{} extraction on BLOOM and mGPT. For the generation-level analysis step, we use BLOOM and GPT2 (focusing on English) expert units and report results for conditioning \cncpt{} datasets in 8 languages: (ara, ben, eng, fra, hin, kor, rus, zho).

\paragraph{Datasets}
As mentioned before, each concept in our dataset contains 100 positive and 300 negative examples. In some cases, we use up-sampling by repeating the example sentences multiple times when we do not have 100 distinct examples mentioning the \cncpt{} name. In total, we prepare 31 \cncpt{} Datasets (22 Country News-Sources, 13 Languages) and extract expert units conditioning over these datasets. We reported the detailed dataset statistics in Appendix \ref{app:dataset} and in Table \ref{tab:ds_main}.

\paragraph{Generative Scheme:} On average we generate \texttt{112,225} sentences for a given model and \cncpt{} Dataset. For 67 \cncpt{} \expt{}, we randomly choose 5 prefix templates; replace those with all 67 country name and generate 5 sentences with the lowest perplexity per \pref{}; thus \texttt{67x5x67x5=112,225} sentences.

\paragraph{Probing Metrics} We analyze both the Geographic Representation Networks (intrinsic/parameter probing) and the generated texts (extrinsic/generation probing) to answer our Research Questions where we utilize the aid of visualization and three additional quantitative metrics as follows:

\noindent \textbf{1. Neighbourhood Score}: We propose a proximity-based metric to quantify the inherent encoding of Geographic Proximity present inside an LM by looking at the country-country associations and compare them with the physical world. For example, in Figure~\ref{fig:main_ex}, South-American neighbouring countries are clustered together thus preserving a factually consistent representation. 
To capture this, we compute the number of neighbours one country node is connected within a 2-hop distance given a \georep{} Network. To better illustrate, consider in a \georep{} Network $G$, country node $c_5 \in G$ is connected with 4 other country nodes $\{c_1,c_2,c_3,c_4\}\in G$. Among these 4 connected nodes, $c_5$ shares sea or land borders with only 2 countries $N_5=\{c_2,c_3\}$ in real world thus making $|N_5|=2$. Similarly, we can compute $|N_2|$ and $|N_3|$ for countries $c_2$ and $c_3$ respectively. So, the Neighbourhood Score $n_s(c_5)=|N_5|+|N_2|+|N_3|$ which we can generalize and aggregate at the network level as follows:
\begin{align*}
    N_s(G) &= \sum_{c_i \in G}n_s(c_i) \\
    & =\sum_{c_i \in G}(|N_i|+\sum_{j \in N_i}|N_j|)
    \end{align*}
\noindent \textbf{2. Representation Score}: We quantify the overall command of prefix, concept or top-represented countries at the \textit{language} level (i.e. for all generated text in a language). Consider we have \expt{} already computed for \cncpt{} $c_i$. We use these units to generate text while providing a \pref{} $p_j$. Later, we map the entities of generated text to countries. So if we have a total of $L=\{l_1,l_2,..l_k..l_n\}$ countries with respective entity counts, we can get the top represented countries $T(c_i,p_j)$ for each concept-prefix pair $(c_i,p_j)$:
    \begin{align*}
    T(c_i,p_j)=\arg\max\limits_{l_k\in L}\left(P(l_k|c_i,p_j)\right)
    \end{align*}
    Having this set of highly represented countries for each concept-prefix pair at hand, we can now compute in how many cases a \cncpt{}, \pref{} or the top-10 most represented countries are present in the set $T(c_i,p_j)$ for all $c_i\in \mathcal{N}$, $p_j\in \mathcal{M} $ where $\mathcal{N}=\{\text{Concept Countries}\}$, $\mathcal{M}=\{\text{Prefix Countries}\}$. So given one output-country-distribution $B$:
    \begin{align*}
    \text{RS}(B,x)= \sum_{c_i \in \mathcal{N}}\sum_{p_j \in \mathcal{M}}|T(c_i,p_j) \in A_x| \ \ 
    \text{where}\\
    A_x= \text{\{prefix $p_j$, concept $c_i$ or top-10 country\}}
    \end{align*}
    The intuition here is to quantify how much the influence of \cncpt{}, \pref{} or overly represented countries varies across languages. For example, if we observe that the score for \pref{} is higher than the scores for \cncpt{} across all settings, it means \pref{} is a more influencing factor than \cncpt{} in the geographical relatedness of the text generation. For comparative analysis, we consider top-3 represented countries instead of just one while computing $T(c_i,p_j) \in A_x$.
    
\noindent \textbf{3. Skewness}\footnote{\url{https://docs.scipy.org/doc/scipy/reference/generated/scipy.stats.skew.html}}: We compare the symmetry of the generated country-entity distribution for both generated and the 
concept dataset texts. The ones that are more skewed one the ones containing amplified bias towards certain country-origin entities.

\section{Findings}
\paragraph{RQ1:} \textit{To what extent the geographic proximity is encoded in the PLMs?}
\begin{figure*}
    \centering
    \begin{tabular}{c}
     Geographical Closeness present in Model Units  \vspace{-.9em}\\\\
    \includegraphics[width=.99\textwidth]{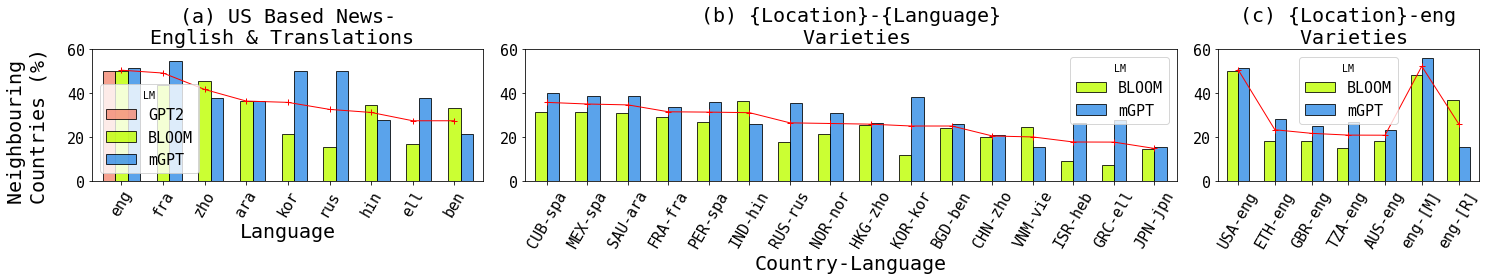}
    \end{tabular}
    \vspace{-1em}
    \caption{(a) The variation of \texttt{neighbourhood score} for different set of expert units. Notice at (a.1) we get the best score for USA-eng and it decreases when we translate the concept dataset. This also varies across languages, models (a.2) and the precise identification of expert units using high-quality concept-dataset also matters (a.3).}
    \label{fig:result_ns}
    \vspace{-1em}
\end{figure*}

\textbf{Intrinsic Findings:} Based on our analysis of the \georep{} Networks, it is evident that model parameters respond similarly for closely related (culturally or geographically) countries. For example, consider the Network in Figure~\ref{fig:main_ex} from BLOOM \expt{} conditioned using the USA-eng \cncpt{} dataset. The Latin-American, African and European blocks are fairly clear. The Indian Subcontinent countries (BGD, PAK, IND), or countries of the British Commonwealth (AUS, NZL, CAN) are also clustered together. In addition, from the communities identified with the Louvain Community Detection algorithm, as visualized in the world map plot, we observe that community clusters are mainly formed around countries with proximity. We prepare similar kinds of \georep{} Networks for all sets of \expt{} conditioned on different \cncpt{} datasets (see Appendix \ref{app:nets}). 

\begin{table}[ht]
    \tiny
    \centering
    \begin{tabular}{l|ll|lll}
    \toprule
     Concept & \multicolumn{2}{c}{Genareted} &\multicolumn{3}{|c}{Expert Units} \\
       & gpt2 & bloom & gpt2 & mgpt & bloom \\
     \midrule
     USA & USA & USA & \rd{SRB} & \rd{SWE} & \rd{SWE} \\
     GBR & GBR & FRA & \rd{POL} & \rd{HUN} & \rd{HUN} \\
     FRA & CHN & IND & BGR & AUT & \rd{SVN} \\
     CHN & IND & GBR & \rd{SVK} & \rd{SVK} & \rd{GRC} \\
     UKR & FRA & CHN & \rd{SWE} & CHN & \rd{SVK} \\
     RUS & CAN & RUS & PER & \rd{GRC} & \rd{POL} \\
     DEU & RUS & JPN & LVA & \rd{POL} & \rd{ARG} \\
     ESP & AUS & KOR & \rd{HUN} & \rd{SVN} & COL \\
     AUS & JPN & DEU & \rd{ARG} & CHL & BRA \\
     JPN & ISR & ESP & TZA & \rd{TUR} & \rd{TUR} \\
    \bottomrule
    \end{tabular}
    \caption{Top represented countries across concepts and generated text. For BLOOM we aggregate across all eight languages; GPT-2 is English only. For expert units, we report the countries with the highest degree of similarity associations.} 
    \label{tab:count_freq}
    \vspace{-2em}
\end{table}

\textbf{Extrinsic Findings:} Next we investigate \textit{whether the encoded geographic proximity gets modified due to geopolitical favouritism} by performing entity-country mapping on a large pool of generated texts in eight languages (112,255 avg. sentences per language). Evidently, we observe a strong presence of \textit{geopolitical favouritism} which we define as the over-amplification of certain country representation (eg. countries with higher GDP, geopolitical stability, military strength etc). For comparison, we use the distribution of the \cncpt{} dataset as it contains the actual news text reflecting real-world affairs. 

\begin{figure*}
    \centering
    \begin{tabular}{c}
     Amplification, Skewness and Representation Bias in Text Generation  \\
    \includegraphics[width=.95\textwidth]{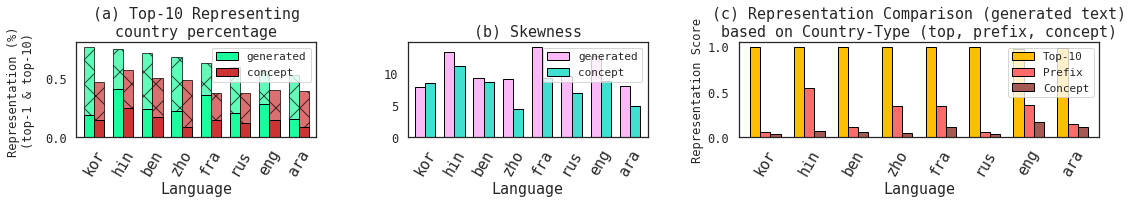}
    \end{tabular}
    \vspace{-1em}    
    \caption{(a) Compared to the concept dataset which is real-world news text, the generated text always overly represents the top-represented countries (eg. USA). (b) This is also true for Skewness (except Korean). In (c) we plot the representation scores depicting the overall influence of prefixes, concepts or top countries. Top countries are over-amplified, irrespective of language. The next dominating factor is prefix but it varies across languages.}
    \label{fig:result_gen}
    \vspace{-1em}
\end{figure*}

In Table~\ref{tab:count_freq} (two left sections), we contrast the top represented countries aggregating the counts from all \cncpt{} datasets to the ones in the generated text. All top-10 most represented countries in generated texts are present within the top-16 ranks of geopolitically significant countries.\footnote{\href{https://worldpopulationreview.com/country-rankings/most-powerful-countries}{worldpopulationreview-powerful-countries}} This resemblance of higher geopolitically powerful country distribution is visible across all forms (Generated text Country Maps in Appendix \ref{app:gen_map}). However, when we compare these top-10 country representations (\%) in generated text with the one from the concept dataset, we observe \textit{geopolitical favouritism}. The result is presented in Figure \ref{fig:result_gen} where in all language country-entity distributions, the top-10 country percentage is always higher compared to real-world news (Figure \ref{fig:result_gen}(a)). A similar pattern is apparent for the other 7 languages (except Korean) in terms of data skewness (Figure \ref{fig:result_gen}(b)). Last, we performed  Kolmogorov–Smirnov and Shapiro statistical significance tests to ensure that the generated text country distribution follows a log-normal distribution. The striking fact here is, though this distribution contains entity mention from 246 countries in total, around \textbf{11.5\%} of all generated entities are from the USA alone. This phenomenon can be further quantified using the neighbourhood score reported in Figure~\ref{fig:result_ns}. For example, as shown in Figure~\ref{fig:result_ns}(a), we find that all 3 models (GPT2, BLOOM, mGPT) \georep{} Networks built from the English dataset conditioned \expt{} have around 50\% of the countries connected with their real-world 2-hop neighbours.

\paragraph{RQ2:} \textit{What is the influence of multilinguality in PLM's knowledge distribution of geographic proximity?}

\textbf{Intrinsic Findings:} By now, we have evidence that Geographic proximity is directly encoded in PLMs in the form of shared expert units. So how this knowledge differs across languages? Ideally, multilingual PLMs should provide equitable utility for their intended users being consistent cross-lingually. To evaluate this, we automatically translate\footnote{Using \url{https://translate.google.com/}} our USA-eng dataset, to avoid any confounders from news content discrepancies from across the world. This way, the content used for identifying the expert units is thematically and semantically the same across languages. The result, in Figure~\ref{fig:result_ns}(a), shows noticeable disparities in \neigh{} percentages across languages in terms of Neighbourhood Scores. When we find \expt{} using Latin-script based \cncpt{} datasets (English, French), the \expt{} make the most of associations among closely related neighbours, while the scores are less than half for Russian, Greek, or Korean in models like mGPT or BLOOM. 

\paragraph{RQ3:} \textit{What is the effect of prompting with geographic identifier (eg. \textit{"In Colombie"} <generate text>) on multilingual text generation?}

\begin{figure}[t]
    \centering
    \includegraphics[width=.5\textwidth]{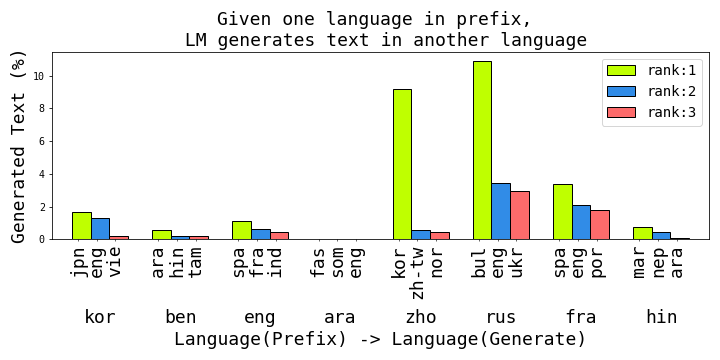}
    \vspace{-1em}
    \caption{Percentage of generated text (top-3) in different language given the Prefix being in another language.}
    \label{fig:lang_id}
    \vspace{-1em}
\end{figure}

\textbf{Extrinsic Findings:} To answer this question, we look into the language of the generated texts using spaCy language identifier\footnote{\href{https://pypi.org/project/spacy-language-detection}{spacy-language-detection}}. On average, BLOOM generates around 5.85\% sentences (52k out of our 898k generated sentences) in a language different than the one of the prefix. This anomaly happens mostly in a larger percentage in Russian, Chinese, and French (Figure \ref{fig:lang_id}). We observe that every language has a specific second language preference (i.e. rank:1 in Figure \ref{fig:lang_id}) which can ignore the given prefix and generate a sentence in that language (eg. kor $\rightarrow$ jap, ben$\rightarrow$ara, eng$\rightarrow$spa, ara$\rightarrow$far, zho$\rightarrow$kor, rus$\rightarrow$bgr, etc). This language preference is not reflexive (eg. kor$\rightarrow$jap whereas zho$\rightarrow$kor). 

Observing the amount of text generated in different languages, it might seem insignificant at first sight. However, we need to keep in mind that there is one geographic identifier in the prefix (\pref{}) as well as given \cncpt{} units. So when we look into which concept-prefix pair usually changes the direction of language, we observe interesting cultural correlations. In Table~\ref{tab:lang_Id}, given a \pref{}, we show how certain country mentions instigate text generation in a different direction (up to 50\% of total generated text, given a prefix-concept pair). This happens frequently when a prefix token is shared among those languages ("in" exists both in English and Spanish; detailed examples in Appendix \ref{app:geo_ind}) and when the country is closely tied with the language. For example, the fra$\rightarrow$spa and eng$\rightarrow$spa directions (French/English prefixes continued in Spanish) include country mentions of Cuba, Argentina, Colombia, or Chile which are all Spanish-speaking countries. We hypothesize that the shared representation space of multilingual decoder often ties language with geographic entity thus changing the favoured generation language.





\begin{table}[t]
    \centering
    \small
\begin{tabular}{@{}l@{ }|@{ }c@{ }c@{ }||l@{ }|@{ }c@{ }c@{}}
    \toprule
    Directions & Concept & Prefix & Directions & Concept & Prefix \\ \midrule
    ben$\rightarrow$ara & LVA & PAK & fra$\rightarrow$spa & CHL & CUB \\ 
    eng$\rightarrow$spa & ARG & COL & fra$\rightarrow$vie & AUT & VNM \\ 
    eng$\rightarrow$ind & IDN & KOR & fra$\rightarrow$por & PRT & PRT \\
    zh-cn$\rightarrow$ko & UGA & NZL & fra$\rightarrow$cat & CHL & SGP \\ 
    rus$\rightarrow$bul & AUS & BGR & fra$\rightarrow$eng & CHL & BGD \\ 
    rus$\rightarrow$eng & ETH & JPN & hin$\rightarrow$mar & BGR & ARE \\ \bottomrule
   \end{tabular}
   \vspace{-1em}
    \caption{Given prefix in one language, the LM generates in a different language, influenced by the concept and prefix countries. These are the cases for which the percentage of language change is more than 50\%.}
    \label{tab:lang_Id}
    \vspace{-2em}
\end{table}

\subsection{Further Analysis}
\paragraph{Data Origin} Because we are experimenting with real-world multilingual news data without going through any extensive data cleaning process, we also need to quantify the dataset-level significance: \textit{how does \cncpt{} data quality impact the identification of \expt{}?} 

The scrapping method we use for dataset construction returns localized news depending on the source location. For example, USA news source provides a higher amount of global news with many country mentions. 
On the other hand, a news source from Bangladesh provides news mostly about its close geopolitical neighbours (eg. India, and China). Thus, the entity frequency distribution of USA-eng and BGD-ben would not be similar. 

In addition, we have variations in the amount of upsampling and the negative instance domain. So in Figures~\ref{fig:result_ns}(b) and~\ref{fig:result_ns}(c), we report \neigh{}s for geographic-source varied on non-English and English datasets respectively. Like before, the association knowledge for USA-eng sourced \georep{} Network remains the most truthful. For Spanish news sourced from different locations (Cuba, Mexico, Peru), scores are rather similar. Interestingly, the score drops significantly for CHN-zho compared to the translated USA-zho from Figure~\ref{fig:result_ns}(a).\footnote{While investigating this anomaly, we found that the fixed sequence length for both models (BLOOM, mGPT) rejects several positive examples during tokenization process thus hurting the \expt{} extraction quality. We corrected this issue by substituting the long examples with shorter ones.}

For the English dataset sourced from different geographic locations (Figure \ref{fig:result_ns}(c)), we get poor association scores for any other locale except the USA, confirming the fact that the in-domain distance between positive and negative examples matters given a fixed language. To dig in further, we perform an ablation study by creating one additional augmented English dataset: eng-[M]: By Masking Country, Name and Organization entities in the USA-eng dataset using Spacy NER. 
Surprisingly, eng-[M] shows the highest percentage of geographic associations even surpassing the original USA-eng one for mGPT. We conclude that small semantic incoherence does not hurt the \expt{} extraction and that more contrastive positive-negative class difference (absence of other entity types) helps.

\paragraph{Model Comparison}
In terms of \neigh{}, mGPT \expt{} encode \textbf{23.5\%} more geographic expertise over BLOOM-560m model on translation datasets (similar text, different language). This improvement is increased \textbf{30\%} when we consider the multilingual datasets (text and language: both different). GPT-2 units perform similarly on the English dataset. 

We conduct another ablation study to quantify how to prune these models towards randomness and semantic incoherence. We prepare another augmented English dataset eng-[R], by putting random semantically incoherent texts while maintaining the positive-negative class difference. The bar showing the \neigh{} is at Figure \ref{fig:result_ns}(c). Now BLOOM \expt{} are almost as good as before, whereas mGPT \expt{} are way worse; only in 3 other cases do BLOOM-560m units represent better associations in total. This reveals that these models contain different distributions even though they were trained with similar objectives, showing different magnitude responses towards data attribute variations, including noise, semantic coherence, data quantity and language.

\paragraph{Influence of \cncpt{} and \pref{}}
We simulate an environment where we provide \expt{} about one geographic entity (\cncpt{}) and ask a PLM about another geographic entity (\pref{}). By now, we have shown that the PLM encodes geographic proximity but also exhibits geopolitical favouritism during inference. The question we ask at this point is: \textit{Given that PLM is biased, how do the \cncpt{} and \pref{} influence text generation?} 

To answer this question, we compute \reps{} on generated texts varying the language (Figure \ref{fig:result_gen}(c)). As always, top-10 country \reps{} is evident in all languages while the second most influencing factor is \pref{}. In Hindi, \cncpt{} has the highest influence of geographic mention in a prompt-based generation. However, this scenario does not hold for the cases of Korean, Bengali, and Russian. On the other hand,  \cncpt{} plays the part of a subtle representative but fails to compete with \pref{} and geopolitical significant countries. One fact to note here is, our experiment contains a small number of examples while generating a large pool of texts. Nevertheless, we believe that it will require intensive data creation efforts to mitigate the biases that coexist with the geographic knowledge in PLMs.

\section{Conclusion and Future Work}
In this study, we perform an experimental analysis on identifying the inherent geographic knowledge and inference bias of prompt-based decoder models. Our experiments strongly suggest that current PLMs are able to encode geographic proximity quiet well. However, almost always geopolitical favouritism overshadows the encoded proximity during inference. This finding raises concerns as well as the need to perform bias-mitigation steps if we want to generate geo-specific texts.  Our additional findings on the impact of multilinguality on prompting points out how encoded geographic proximity is unevenly distributed across languages and how even just a mention of geographic identifiers may influence the language of free-form text generation.

We believe these findings still leave issues to be addressed in current practice and that there should be a a fundamental multilingual-bias mitigation step included in any NLP task workflow. Keeping this in mind, we want to expand the domain of our proposed probing framework and assess its applicability beyond geography. In addition, we aim to perform contrastive training to efficiently extract expert units thus stepping forward with the effort of reducing the inequality inherent in multilingual language models.

\section*{Limitations}
First of all, selecting country as geographic entities is inherently lossy and ideally, we would be able to perform the experiments with further granularity. We rely on Wikidata for entity linking, which is already somewhat biased towards western countries. In addition, our experiments are limited to 69 countries and 13 languages (8 for generating text) (by necessity and due to computing costs), ignoring other countries as well as languages, especially low-resource ones. In the future, we want to further expand our study to include a lot more languages and cultures.

\section*{Acknowledgements} This work is generously supported by the National Science Foundation under grants FAI-2040926 and IIS-2127901.

\bibliography{anthology,custom}
\bibliographystyle{acl_natbib}

\clearpage
\newpage

\appendix
\onecolumn
\section{Terminologies} 
Based on our Framework description, let us list some terminologies that we use for the remainder of the paper, to describe the experimental settings and results.
\begin{enumerate}[nolistsep,noitemsep,leftmargin=*]
\item \textbf{\cncpt{}}: These are the countries for which we collect news.
 \item \textbf{Source Country}: These are the countries \textit{from} which the news data comes from.
    \item \textbf{\cncpt{} Dataset}: A \cncpt{} dataset named \textit{C-l} contains news from \textit{Source Country} \textit{C} in language \textit{l}. It contains concept set $CS=\{c_i,c_j,..c_n\}$ where $c_i$ is one \cncpt{}. Each $c_i$ has 100 positive examples (mention of $c_i$) sentences and 300 negative examples (no mention) sentences.
    \item \textbf{Prefix}:  This is the text that we use to prompt the model, which may include a country mention. This country is the \pref{}.
    \item \textbf{Language}: The language that both the concept dataset and the generated text are in.
    \item \textbf{\expt{}}: The units that are specific to a country concept $c_i$ and are extracted from the language models. 
\end{enumerate}   

\section{Frequently asked questions}
\subsection{What does it mean by the term geographic biases, geographic favouritism and what are their relationships with fairness?}
In general, geographic bias means the over-representation of certain geographic attributes. In this study, we use \textit{"geographic bias"} and \textit{"geographic favouritism"} interchangeably as the  over-amplification of certain country representation (eg. countries with higher GDP, geopolitical stability, military strength etc) during PLM prediction or text-generation. We believe the overall system utility of a language model should be equitable according to the needs of the intended users with different demographic and geographic origin. Thus ensuring their geographic characteristics are well-represented and not over-shadowed because of geographic favouritism is defined as \textit{"geographic fairness"} in this study.

\subsection{What's the reason for using the self-conditioning approach of \citet{suau2022selfcond} for studying biases? There had been many other bias measures in NLP before \citet{suau2022selfcond}. Are they not suitable for the study of geographic and geopolitical biases?
}
A number of previous studies experimented with the behavior different PLMs exhibits while probing with geographic-context as well as cultural-commonsense   \cite{yin2022geomlama, ghosh-etal-2021-detecting}. 
However, we need to extract the specific model weights responsible for these observable polarity. Then using those weights in a controlled setting, we might be able to unfold how PLMs encode geographic knowledge as well as explain the exhibition of geographic-bias during inference. The self-conditioning model proposed by \citet{suau2022selfcond} is one such study that fits to our intended needs perfectly. This approach serves two main purposes: (1) Identifying expert units: model parameters responsible for generating text related to the target concept (i.e. doctor). (2) Triggering specific behaviour in text generation without explicit mentioning or fine-tuning of the target context, which inadvertently influences the behaviour of the model utilizing the encoded-knowledge of PLM.

\subsection{What are the practical takeaways from this? Yes, different models encode geographic knowledge, so what? Should we be concerned, should we do something about it?
}
 We recall the example presented earlier: consider a $L_1$ Spanish speaker from Peru, who is using a prompt-based PLM (like that of~\citet{wang-etal-2022-promda, https://doi.org/10.48550/arxiv.2109.09193}) to generate a localized synthetic dataset for some downstream task. They may use Spanish \textit{as used in the local context} to form their seed data/prefix/prompts. Now, if this language model has already skewed preferences towards geopolitically important countries, it is likely the generated texts will reflect this skewness, thus not appropriately reflecting the local, Peruvian context that the practitioner is interested in. In this study we address this concern of geographic bias being one of the most-significant yet ignored attributes in practice. Moreover, we show how this is further amplified when we go beyond English and similar languages. Basically we need effective bias-mitigation module as part of the regular NLP workflow which is currently non-existent.   

\subsection{Why we need to extract the \expt{} and how \cncpt{} helps in this regard?}
One of our aims is to unfold the geographic representation using relevant PLM units without external fine-tuning. So, we need to find or extract these relevent units which are basically model parameters. So, we can use our \cncpt{} datasets as binary classification dataset (positive class contains sentences mentioning certain\cncpt{}) to find these highly responsive weights (i.e. \expt{}) to certain \cncpt{}. Then we perform self-conditioning on the PLMs using these \expt{} to generate texts having the influence of these \cncpt{}s. 

\subsection{Explain \cncpt{} dataset creation process.}
We scrape news using a Google news api\footnote{\url{https://github.com/ranahaani/GNews}} to capture the current affairs. Importantly, we can select news not just from a given date range, but also news originating in a specific country and a specific language. Such a dataset should allow us to get a reasonable representation of current geopolitical affairs. As such, each of the concept datasets we create reflects \textit{``current news about a country reported by the mainstream platforms from another country"}. Hence, a \cncpt{} dataset \bc{\texttt{\{C\}-\{l\}}} contains news about several (${c_1,c_2,..c_n}$) countries in \texttt{\{l\}} language where the news-source is \texttt{\{C\}} country. For example, \texttt{USA-eng} contains data from US sources, in English, which either mention other countries (there are 100 positive examples for each country $c_i$) or are random sentences not mentioning any countries (negative examples). 

\subsection{Explain the \expt{} extraction process.}
Consider the \cncpt{} \texttt{India} from the dataset \texttt{USA-eng}. Essentially, we have positive examples (text mentioning India or relevant entities) and negative examples (random other sentences not mentioning India)
which we can use to identify the model's \expt{}. These units are the neurons which can be used as predictors to identify the presence of a concept (i.e. positive examples mentioning "India"). The self-conditioning framework computes these neurons and uses the average-precision score to rank their predictive expertise thus allowing us to select the top-$k$ (eg. 10, 50) \expt{} from each layer. 

\subsection{What does Geographic Representation Network actually represents?}
Note that these networks are produced using the uncovered original PLM expert units, without any external data fine-tuning or prompting. Hence, they provide a view of the \textit{inherent} geographic knowledge present inside the PLM parameter space.

\subsection{Why we need to use \expt{} during text generation?}
We have a setting where we can provide certain \cncpt{} as part of the generation condition and the specific \expt{} from the model itself are supposed to be capable enough to influence the generated text. Our aim is to evaluate the geographic knowledge specific model weights or \expt{} by asking those about other \pref{}. This will unfold whether the geopolitical favouritism happens for geopolitically important countries or the geographical proximity (eg. neighbouring countries) takes the precedence or there exist no such patterns.

\subsection{What are the factors considered while constructing the \cncpt{} dataset?}
There are two relevant factors: (1) For the negative examples in USA-eng \cncpt{} dataset, we use news from a completely different domain (eg. automobile, sport), whereas for different geographic-sourced datasets, negative examples come from randomly sampling news of different locations. (2) The intensity of text-noise and positive example up-sampling amount varies across different news-sourced \cncpt{} datasets.

\subsection{Why 2-hop distance while calculating the neighbourhood-score?
}
We did experiment with n-hop scoring and they follow similar trends. We choose 2-hop is it is less complex for scoring and at the same-time, sufficient to point out the disparity across multiple languages.

\subsection{Comparison to news: although these models are trained on web text, which contain news articles, they are not guaranteed to generate text like a news article. Thus the distribution of entities within the text will be different.
}
Yes, that is correct but our aim is to capture the learned distribution and evaluate (1) whether that distribution is skewed or not, (2) Whether there is resemblance with the real-world scenario or not. We believe, this assessment is important for a PLM which will be used for solving real-world practical tasks and having news-text for comparison might be the closest viable source we can get in a limited resource setting.

\section{Datasets}
\label{app:dataset}

In Table \ref{tab:ds_main} we present the concept dataset details. Each dataset here contains 43 to 69 country concept files (The complete list of countries is presented in Table \ref{tab:clist}). The \texttt{Type-2} datasets are the translated version of USA-eng dataset. In \texttt{Type-3}, we mask USA-eng entities using a NER tagger and \texttt{Type-4} is constructed using random english texts.


\begin{table*}
\centering

\begin{tabular}{p{5.5cm}|p{.4cm}|p{7cm}@{}}
\toprule
 \textbf{Dataset Names}  & \# & Description \\ \midrule
\multicolumn{3}{c}{\textbf{\texttt{Type 1: \{News\_Source\_Location\}-\{Language\}}}} \\
\vspace{.04em}
\begin{tabular}{l|l|l}
   \uline{\textit{USA-eng}} & \uline{\textit{BGD-ben}} & \uline{\textit{CHN-zho}}  \\
    GRC-ell& ISR-heb& \uline{\textit{IND-hin}}\\
    \uline{\textit{KOR-kor}}& MEX-spa& NOR-nor\\
    \uline{\textit{SAU-ara}}& VNM-vie& AUS-eng\\
    ETH-eng& GBR-eng& HKG-zho\\
    TZA-eng & \uline{\textit{FRA-fra}} & PER-spa\\
    JPN-jpn & \uline{\textit{RUS-rus}} & CUB-spa\\
\end{tabular} & 
\multirow{10}{*}{21} &
\vspace{.04em}
These 21 datasets are scrapped from news sources originating from 21 different countries in different languages. Each one of these datasets contain country concept sets describing news about specific countries. Each country concept are prepared using 100 positive sentence examples and 300 negative sentence examples. We use upsampling by repetition when we have less examples than the required counts. For only \texttt{USA-eng} dataset, we use english news from other topic search (eg. \textit{Automotive, Sport}) to construct the negative examples while, for other 20 datasets we use news about other countries (i.e. in domain) as negative examples.\\
\midrule
\multicolumn{3}{c}{\textbf{\texttt{Type 2: \{News\_Source\_USA\}-\{Translations\}}}} \\
\vspace{.01em}
\begin{tabular}{l|l|l}
    USA-ara & USA-ben& USA-ell \\
    USA-hin& USA-kor& USA-rus\\
    USA-zho & USA-fra
\end{tabular} & 
\multirow{4}{*}{8} &
\vspace{.04em}
These 8 datasets are created using translation from the USA-eng dataset. We use Google Translation API$^1$  to translate the texts from source language to target language.\\
\midrule
\multicolumn{3}{c}{\textbf{\texttt{Type 3: \{USA-eng\}-\{Masked Entities\}}}} \\
\vspace{.01em}
\begin{tabular}{llll}
    USA-eng-[M] \\
\end{tabular} & 
\multirow{3}{*}{1} &
\vspace{.04em}  
We augment \texttt{USA-eng} dataset by masking all additional entities in positive examples for each country concepts using spaCy$^2$.\\
\midrule
\multicolumn{3}{c}{\textbf{\texttt{Type 4: \{USA-eng\}-\{Random Text\}}}} \\
\vspace{.01em}
\begin{tabular}{llll}
    eng-[R] \\
\end{tabular} & 
\multirow{4}{*}{1} &
\vspace{.04em}  
We randomly use text instead of original text in \texttt{USA-eng} dataset while maintaining the positive negative class distinction but without any semantic coherence.\\
\bottomrule
 \multicolumn{3}{l}{\footnotesize{[1] \url{https://translate.google.com/}}} \\
 \multicolumn{3}{l}{\footnotesize{[2] \url{https://spacy.io/}}} \\
\end{tabular}

\caption{Country Concept Datasets sourced from Google News texts. We extracted expert units from language models: gpt-2 (only english), bloom and mgpt for all of these. Among these, we perform text generation using the expert units sourced from \uline{\textit{8 datasets}} (The underline ones).}
\label{tab:ds_main}
\vspace{-1em}
\end{table*}

\begin{table*}[t]
\centering
\begin{tabular}{ll|ll|ll}
\toprule
   ISO &              Country &    ISO &                    Country &    ISO &                 Country \\
\midrule
 AUS &      Australia &  BWA &             Botswana &  CAN &           Canada  \\
 ETH &       Ethiopia &  GHA &                Ghana &  IND &            India  \\
 IDN &      Indonesia &  IRL &              Ireland &  ISR &           Israel  \\
 KEN &          Kenya &  LVA &               Latvia &  MYS &          Malaysia \\
 NAM &        Namibia &  NZL &          New Zealand &  NGA &           Nigeria \\
 PAK &       Pakistan &  PHL &          Philippines &  SGP &         Singapore \\
 ZAF &   South Africa &  TZA &             Tanzania &  UGA &            Uganda \\
 GBR & United Kingdom &  USA &        United States &  ZWE &          Zimbabwe \\
 CZE & Czech Republic &  DEU &              Germany &  AUT &           Austria \\
 CHE &    Switzerland &  ARG &            Argentina &  CHL &             Chile \\
 COL &       Colombia &  CUB &                 Cuba &  MEX &            Mexico \\
 PER &           Peru &  VEN &            Venezuela &  BEL &          Belgium  \\
 FRA &         France &  MAR &              Morocco &  SEN &           Senegal \\
 ITA &          Italy &  LTU &            Lithuania &  HUN &           Hungary \\
 NLD &    Netherlands &  NOR &               Norway &  POL &            Poland \\
 BRA &         Brazil &  PRT &             Portugal &  ROU &           Romania \\
 SVK &       Slovakia &  SVN &             Slovenia &  SWE &            Sweden \\
 VNM &        Vietnam &  TUR &               Turkey &  GRC &            Greece \\
 BGR &       Bulgaria &  RUS &               Russia &  UKR &          Ukraine  \\
 SRB &         Serbia &  ARE & United Arab Emirates &  SAU &      Saudi Arabia \\
 LBN &        Lebanon &  EGY &                Egypt &  BGD &        Bangladesh \\
 THA &       Thailand &  CHN &                China &  TWN &            Taiwan \\
 HKG &      Hong Kong &  JPN &                Japan &  KOR & Republic of Korea \\
None &           None & None &                 None & None &              None \\
\bottomrule
\end{tabular}
    \caption{List of Countries we conducted experiments on.}
    \label{tab:clist}
\end{table*}

\section{Prefix Templates}
For each of the eight languages, we generate prefix replacing templates with \pref{} names. Per language, we have six template prefix. The complete list is presented in Table \ref{tab:template}
\label{app:prefix}
\begin{table*}[]
    \centering
    \begin{tabular}{c}
         \includegraphics[width=.8\textwidth]{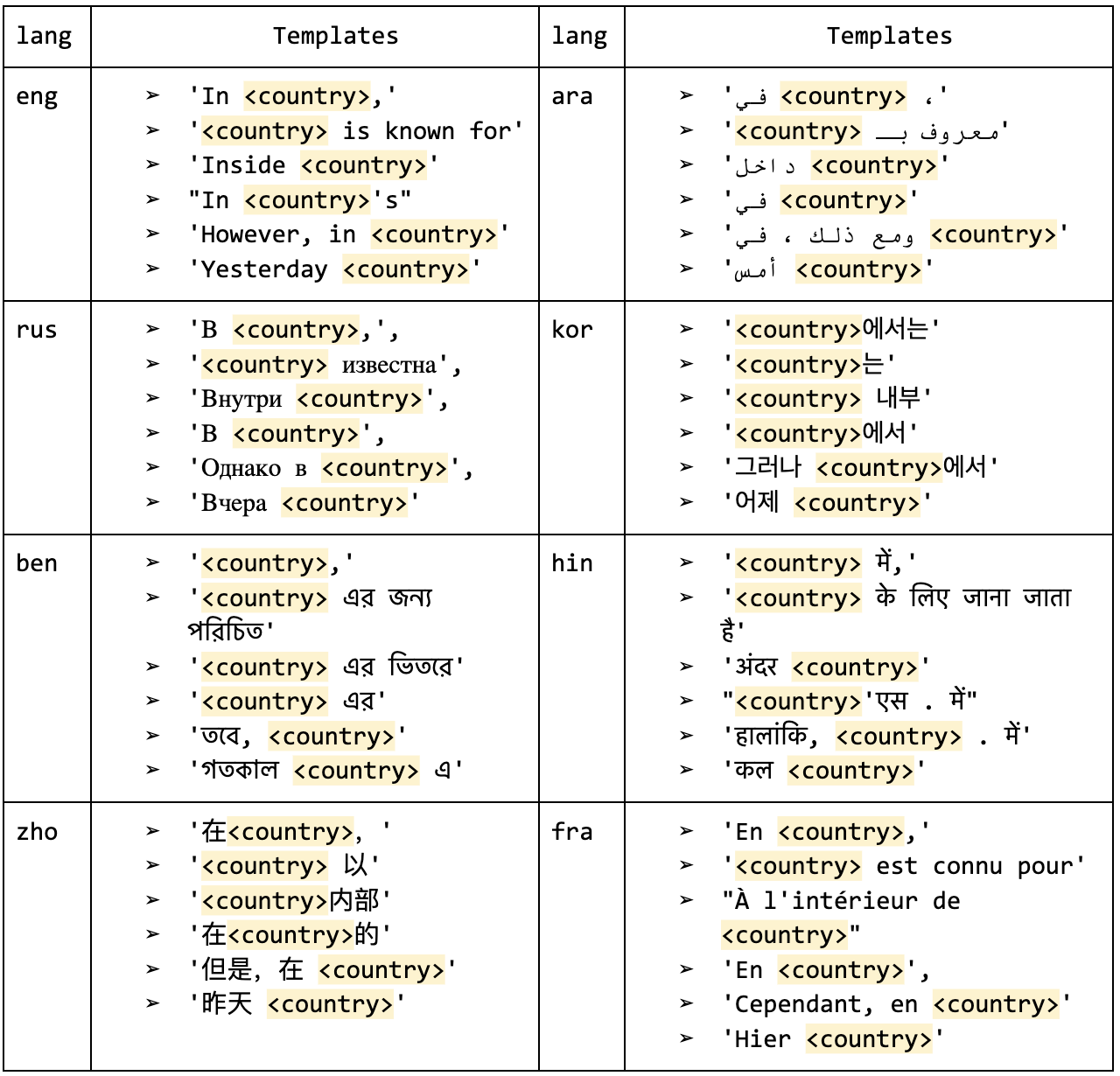}
    \end{tabular}
    \caption{Prefix templates we use for Multilingual Text Generation. We replace the \textbf{\texttt{<country>}} with the corresponding country name in generator language. For example, To construct one USA-mention Chinese prefix, we replace \textbf{\texttt{<country>}} with \textbf{\texttt{\begin{CJK*}{UTF8}{gbsn}美国\end{CJK*}}}.  We use a multilingual country-name dataset~\cite{cname} to query country names.}
    \label{tab:template}
\end{table*}

\section{Additional Geographic Representation  Networks}
\label{app:nets}
\begin{figure*}[t]
    \centering
    \begin{tabular}{cc}
    \multicolumn{2}{c}{\textbf{Geographic Representation Networks and Corresponding Community Maps}}\\\\
        \includegraphics[width=.45\textwidth]{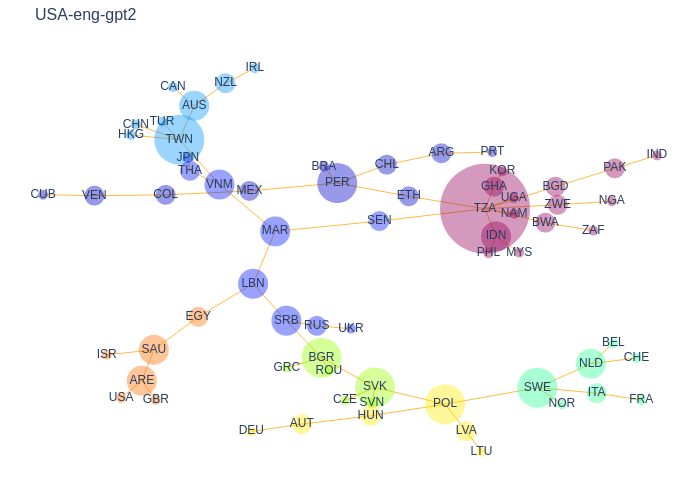}
        & \hspace{-4em}\includegraphics[width=.45\textwidth]{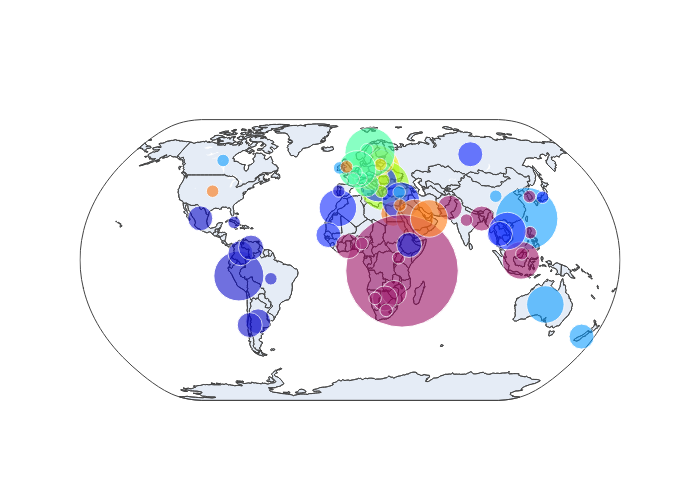}
        \vspace{-1.5em} \\ 
        \multicolumn{2}{c}{(1)}\\
        \includegraphics[width=.45\textwidth]{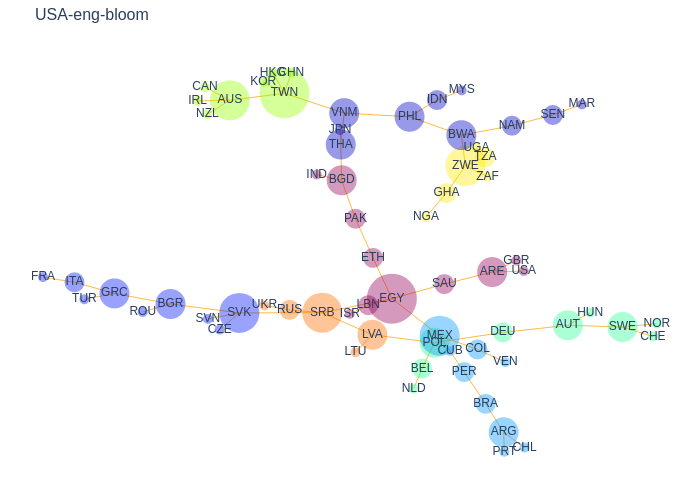}
        & \hspace{-4em}\includegraphics[width=.45\textwidth]{plot/USA-eng-bloom-csim-c.png}
        \vspace{-1.5em} \\ 
        \multicolumn{2}{c}{(2)}\\
        \includegraphics[width=.45\textwidth]{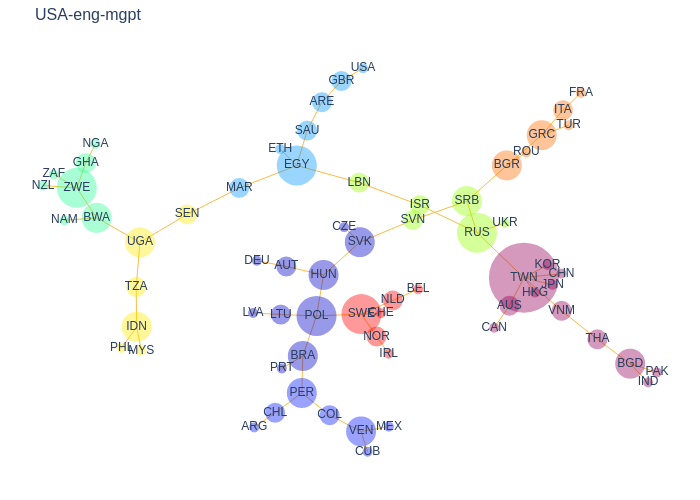}
        & \hspace{-4em}\includegraphics[width=.45\textwidth]{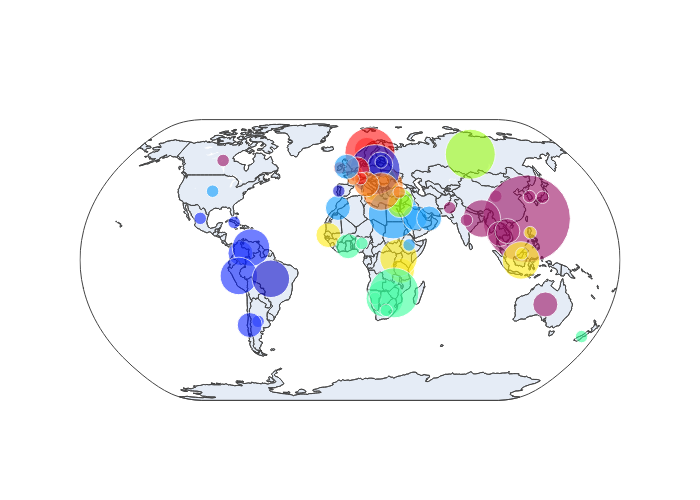}
        \vspace{-1.5em} \\ 
        \multicolumn{2}{c}{(3)}\\
        \includegraphics[width=.45\textwidth]{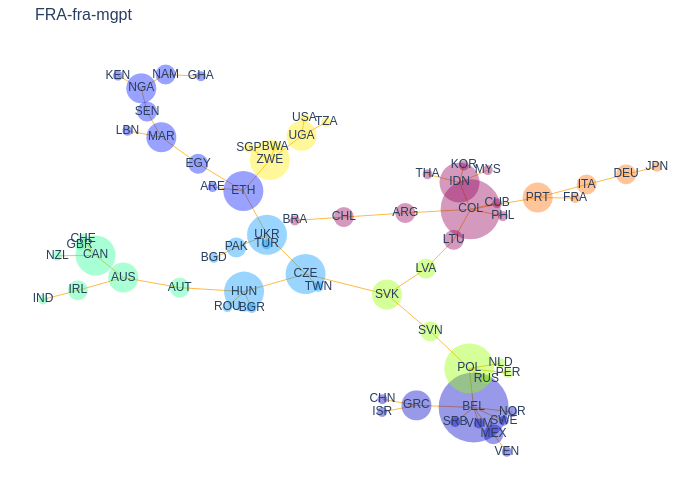}
        & \hspace{-4em}\includegraphics[width=.45\textwidth]{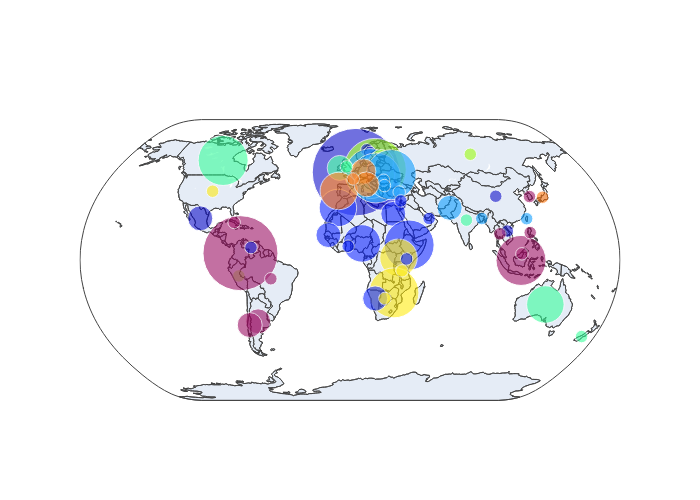}
        \vspace{-1.5em} \\ 
        \multicolumn{2}{c}{(4)}\\
    \end{tabular}
    \caption{Geographic Representation Network and Corresponding Community Map for different Expert Unit set Associations. The language models we use are GPT2 (only English), mGPT and BLOOM.}
    \label{fig:plot_exp_1}
\end{figure*}

\begin{figure*}[t]
    \centering
    \begin{tabular}{cc}
        \includegraphics[width=.45\textwidth]{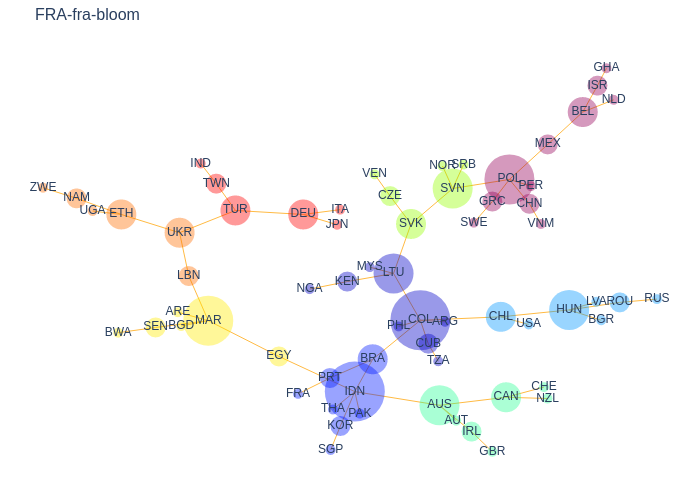}
        & \hspace{-4em}\includegraphics[width=.45\textwidth]{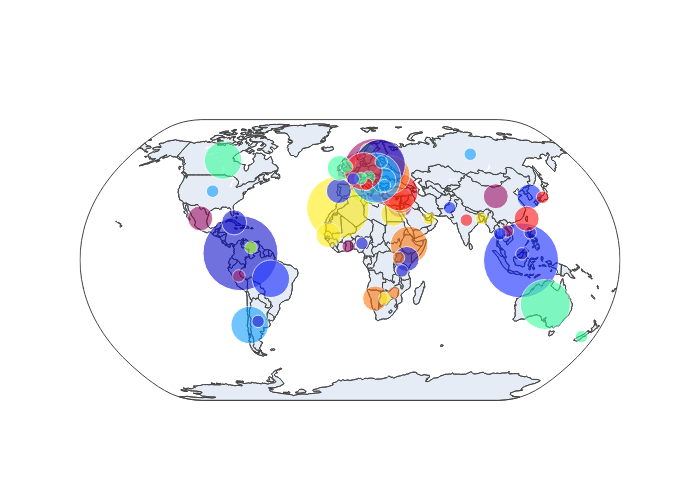}
        \vspace{-1.5em} \\ 
        \multicolumn{2}{c}{(5)}\\
        \includegraphics[width=.45\textwidth]{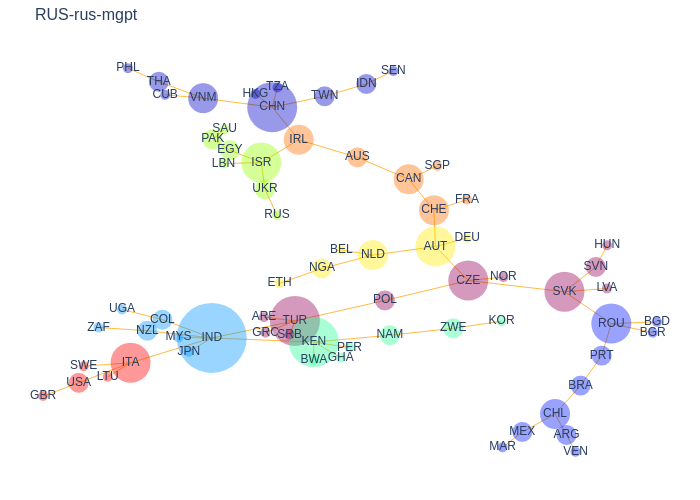}
        & \hspace{-4em}\includegraphics[width=.45\textwidth]{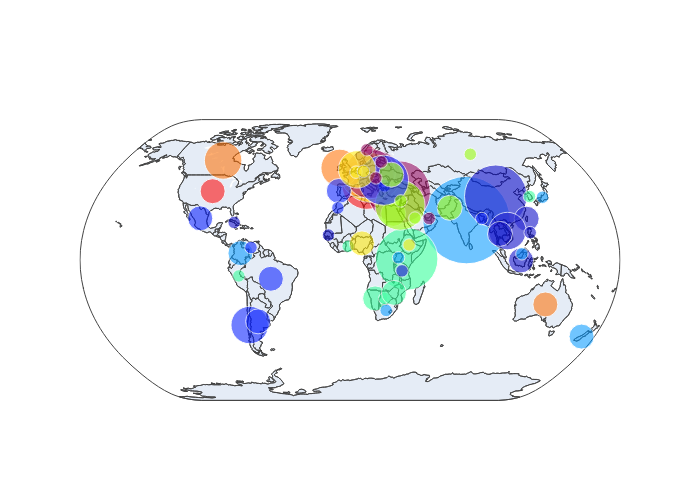}
        \vspace{-1.5em} \\ 
        \multicolumn{2}{c}{(6)}\\
        \includegraphics[width=.45\textwidth]{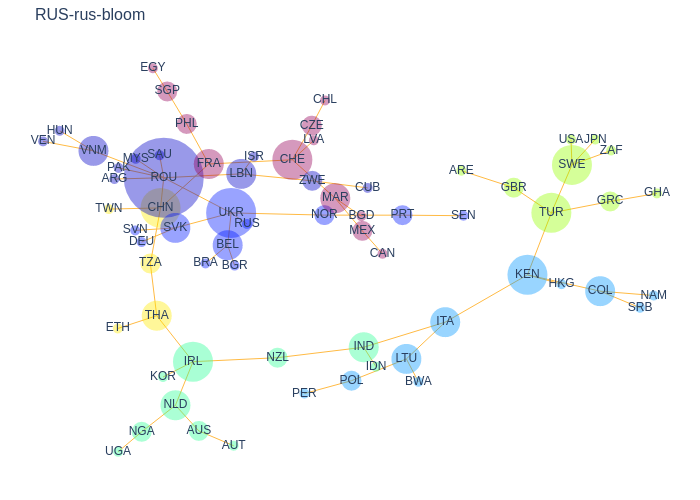}
        & \hspace{-4em}\includegraphics[width=.45\textwidth]{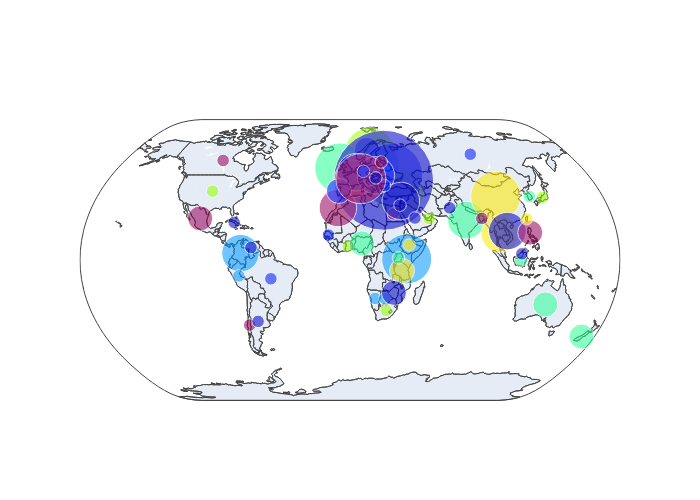}
        \vspace{-1.5em} \\ 
        \multicolumn{2}{c}{(7)}\\
        \includegraphics[width=.45\textwidth]{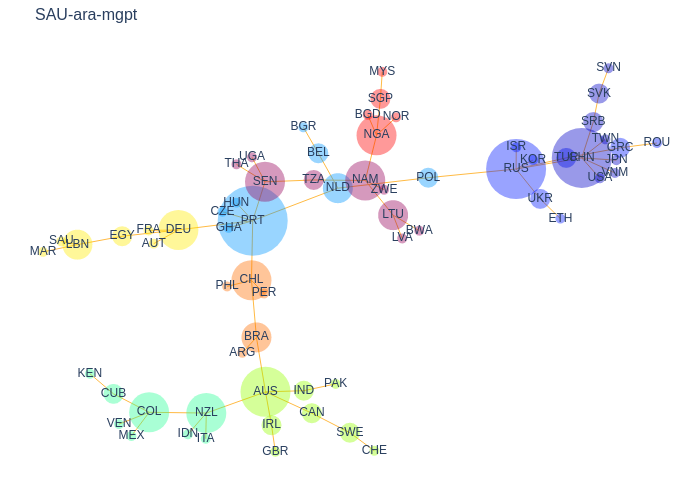}
        & \hspace{-4em}\includegraphics[width=.45\textwidth]{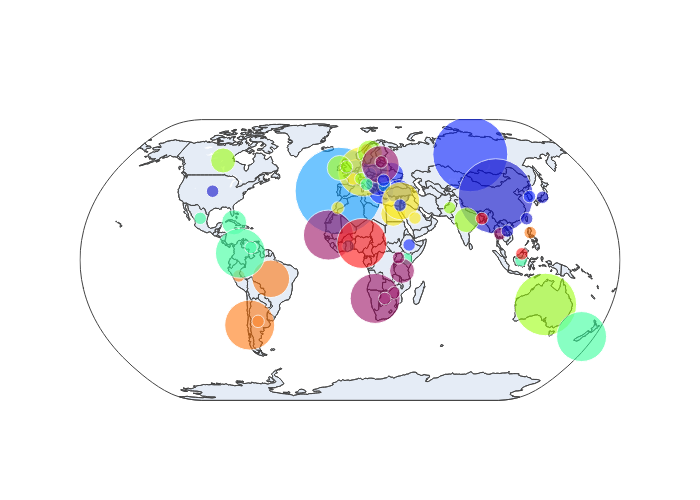}
        \vspace{-1.5em} \\ 
        \multicolumn{2}{c}{(8)}\\
    \end{tabular}
    \caption{Geographic Representation Network and Corresponding Community Map for different Expert Unit set Associations. The language models we use are GPT2 (only English), mGPT and BLOOM.}
    \label{fig:plot_exp_2}
\end{figure*}

\begin{figure*}[t]
    \centering
    \begin{tabular}{cc}
        \includegraphics[width=.45\textwidth]{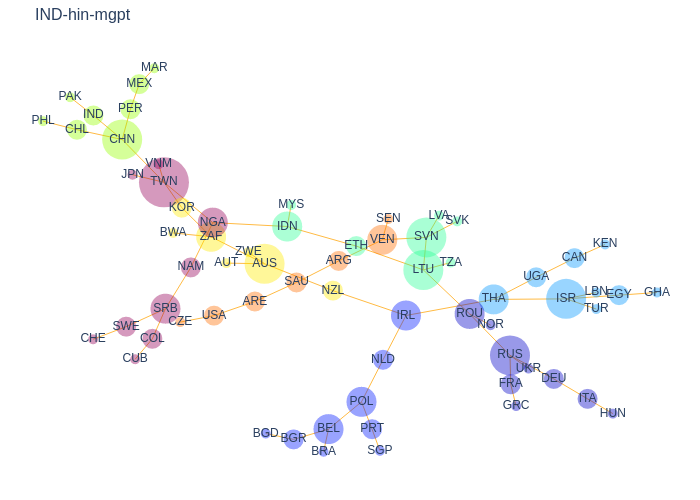}
        & \hspace{-4em}\includegraphics[width=.45\textwidth]{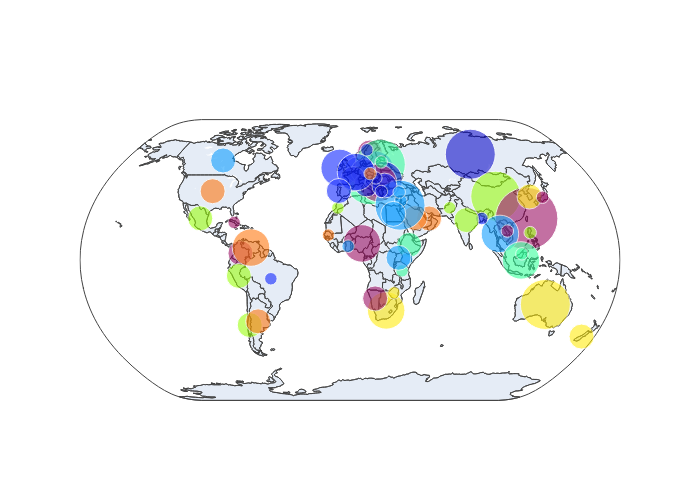}
        \vspace{-1.5em} \\ 
        \multicolumn{2}{c}{(9)}\\
        \includegraphics[width=.45\textwidth]{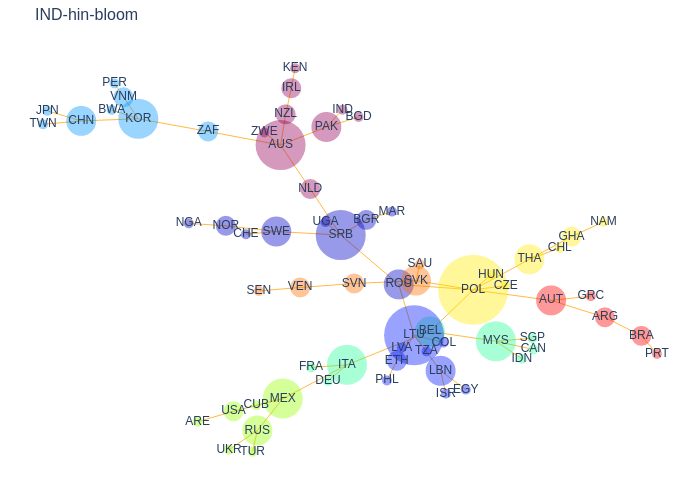}
        & \hspace{-4em}\includegraphics[width=.45\textwidth]{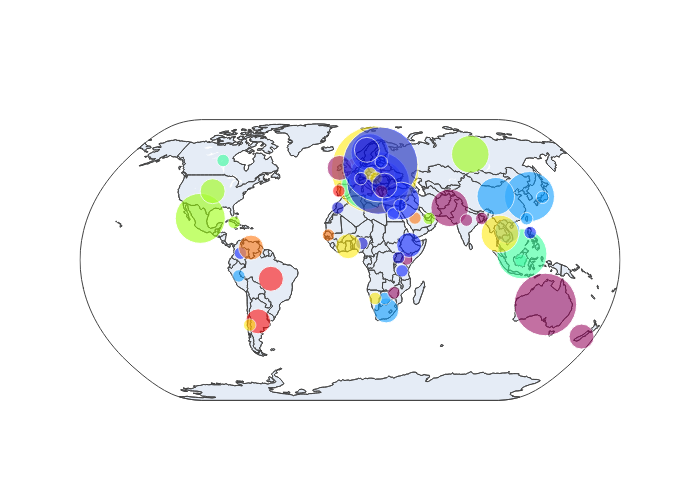}
        \vspace{-1.5em} \\ 
        \multicolumn{2}{c}{(10)}\\
        \includegraphics[width=.45\textwidth]{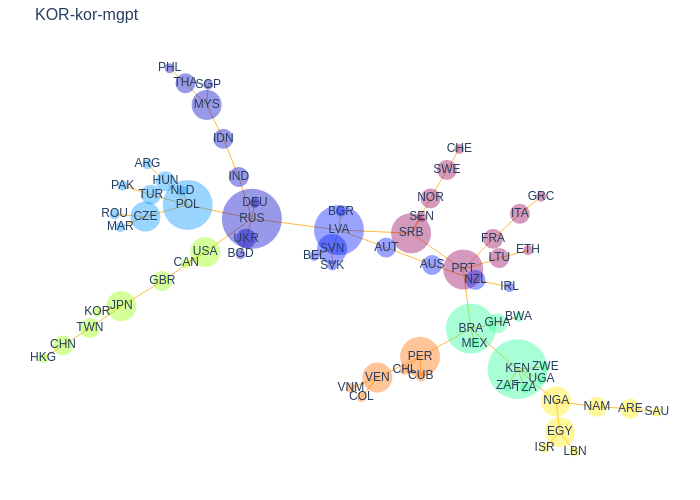}
        & \hspace{-4em}\includegraphics[width=.45\textwidth]{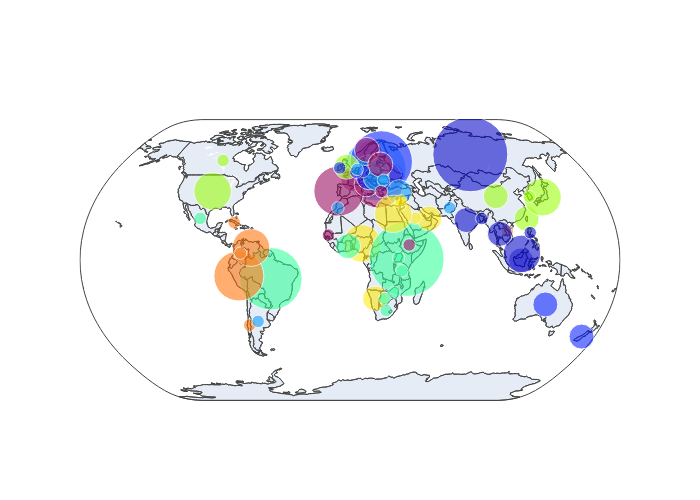}
        \vspace{-1.5em} \\ 
        \multicolumn{2}{c}{(11)}\\
        \includegraphics[width=.45\textwidth]{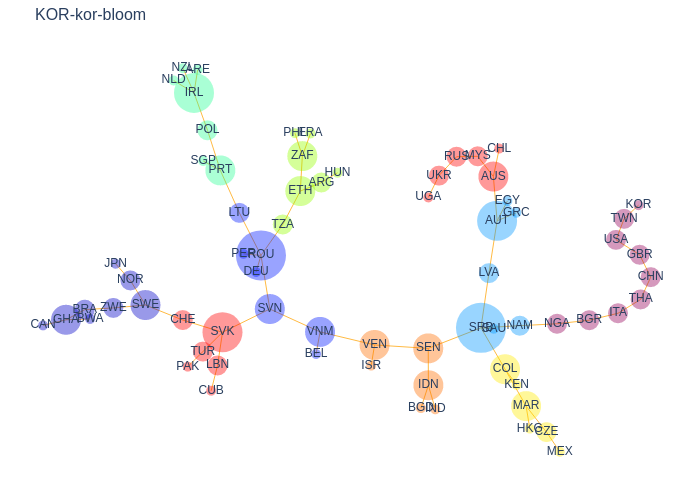}
        & \hspace{-4em}\includegraphics[width=.45\textwidth]{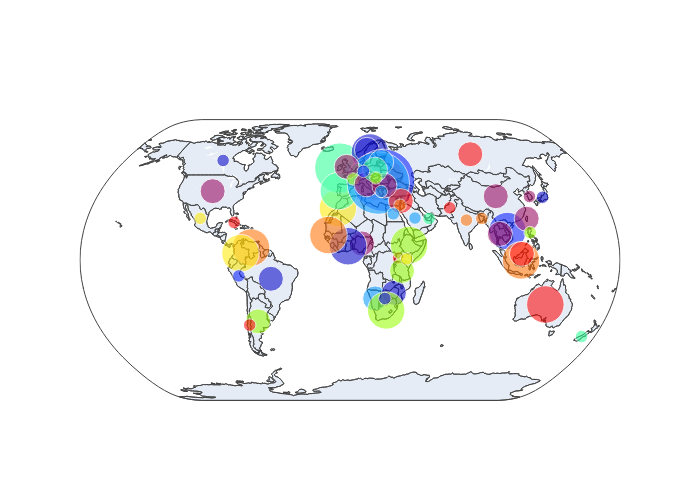}
        \vspace{-1.5em} \\ 
        \multicolumn{2}{c}{(12)}\\
    \end{tabular}
    \caption{Geographic Representation Network and Corresponding Community Map for different Expert Unit set Associations. The language models we use are GPT2 (only English), mGPT and BLOOM.}
    \label{fig:plot_exp_3}
\end{figure*}

\begin{figure*}[t]
    \centering
    \begin{tabular}{cc}
        \includegraphics[width=.45\textwidth]{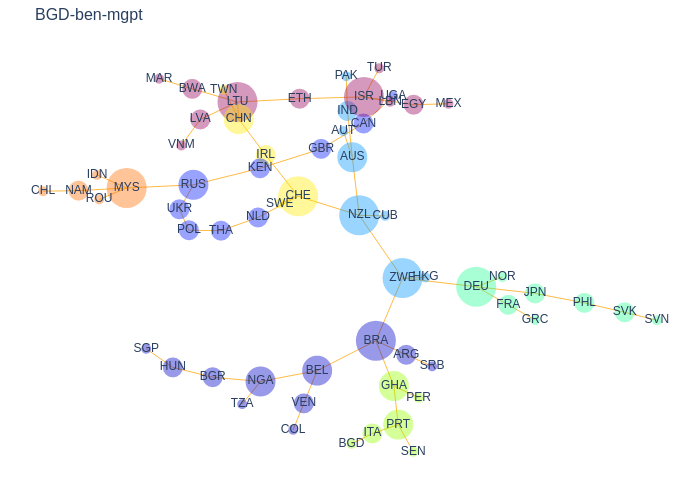}
        & \hspace{-4em}\includegraphics[width=.45\textwidth]{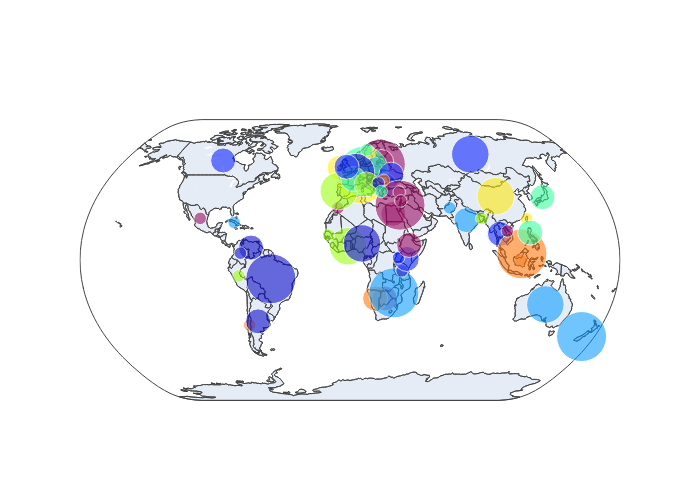}
        \vspace{-1.5em} \\ 
        \multicolumn{2}{c}{(13)}\\
        \includegraphics[width=.45\textwidth]{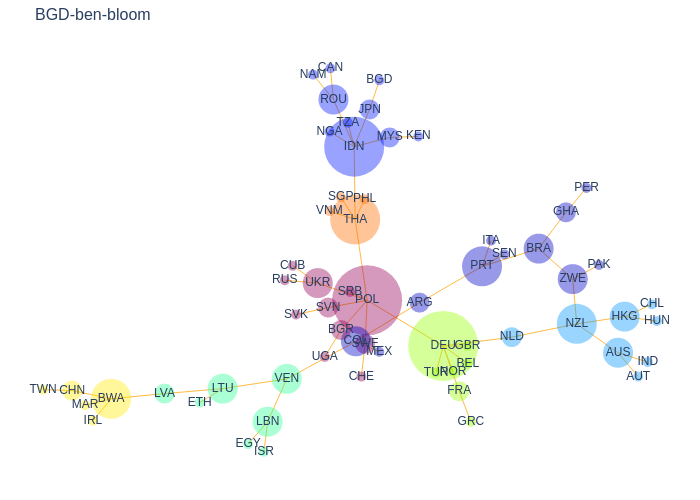}
        & \hspace{-4em}\includegraphics[width=.45\textwidth]{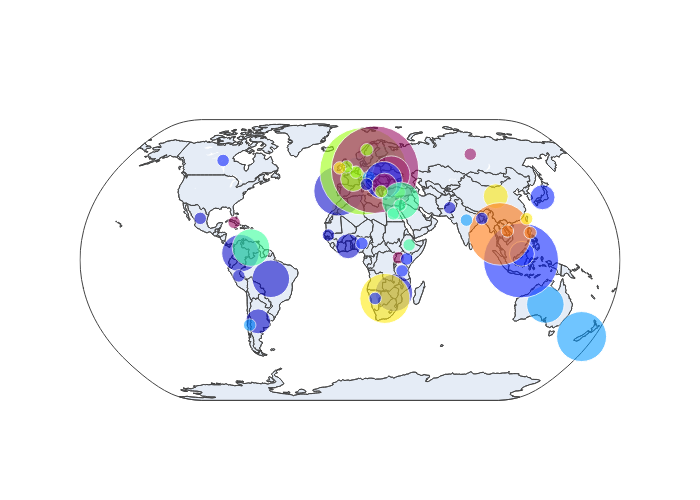}
        \vspace{-1.5em} \\ 
        \multicolumn{2}{c}{(14)}\\
        \includegraphics[width=.45\textwidth]{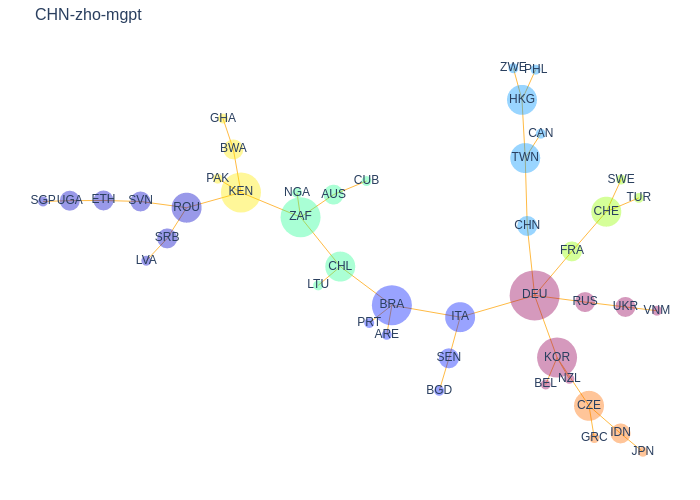}
        & \hspace{-4em}\includegraphics[width=.45\textwidth]{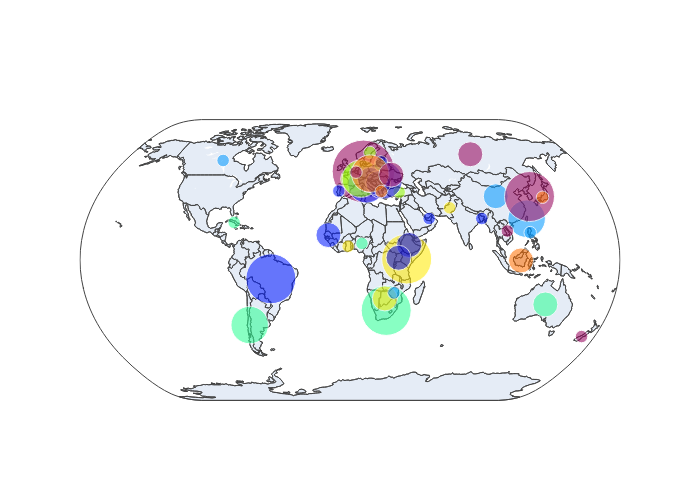}
        \vspace{-1.5em} \\ 
        \multicolumn{2}{c}{(15)}\\
        \includegraphics[width=.45\textwidth]{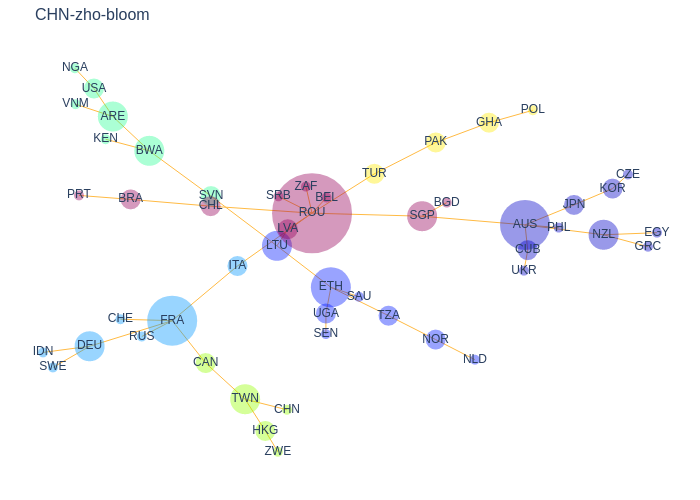}
        & \hspace{-4em}\includegraphics[width=.45\textwidth]{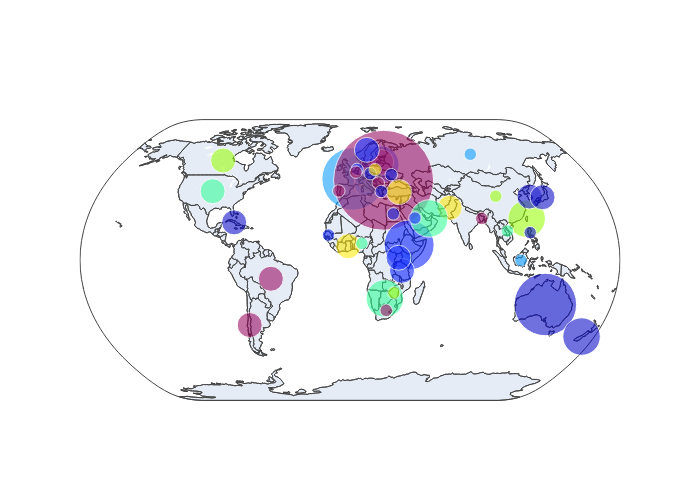}
        \vspace{-1.5em} \\ 
        \multicolumn{2}{c}{(16)}\\
    \end{tabular}
    \caption{Geographic Representation Network and Corresponding Community Map for different Expert Unit set Associations. The language models we use are GPT2 (only English), mGPT and BLOOM.}
    \label{fig:plot_exp_4}
\end{figure*}
In Figures (\ref{fig:plot_exp_1}, \ref{fig:plot_exp_2}, \ref{fig:plot_exp_3}, \ref{fig:plot_exp_4}) we present \georep{} Networks (News Source-language: USA-eng, SAU-ara, FRA-fra, RUS-rus, BGD-ben, KOR-kor, CHN-zho, IND-hin) constructed using the \expt{} from GPT2, BLOOM and mGPT. 

\section{Geography Maps on generated text}
\label{app:gen_map}
\begin{figure*}[t]
    \centering
    \begin{tabular}{cc}
    \multicolumn{2}{c}{\textbf{Geographic Representation Networks and Corresponding Community Maps}}\\\\
        \includegraphics[width=.5\textwidth]{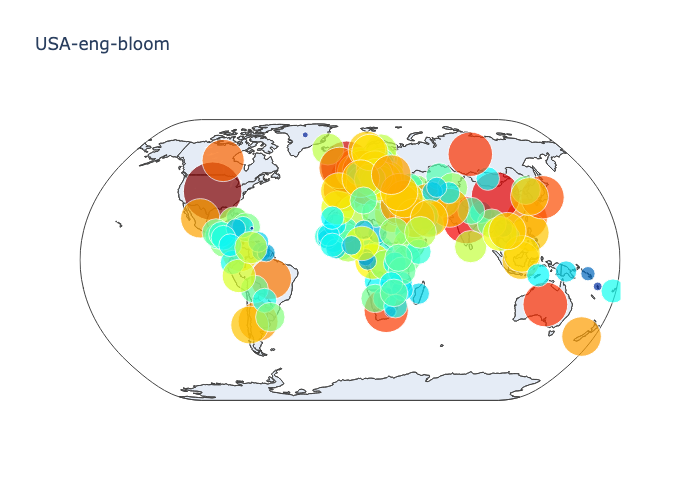}
        & \hspace{-4em}\includegraphics[width=.5\textwidth]{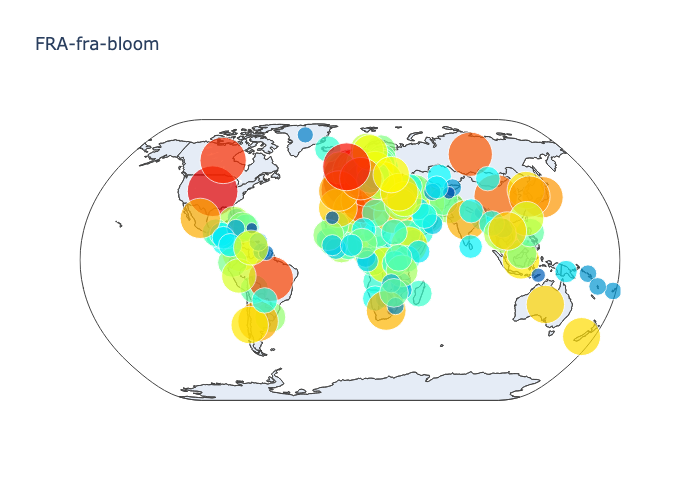}
        \vspace{-2.5em} \\ (a) & (b)\\
        \includegraphics[width=.5\textwidth]{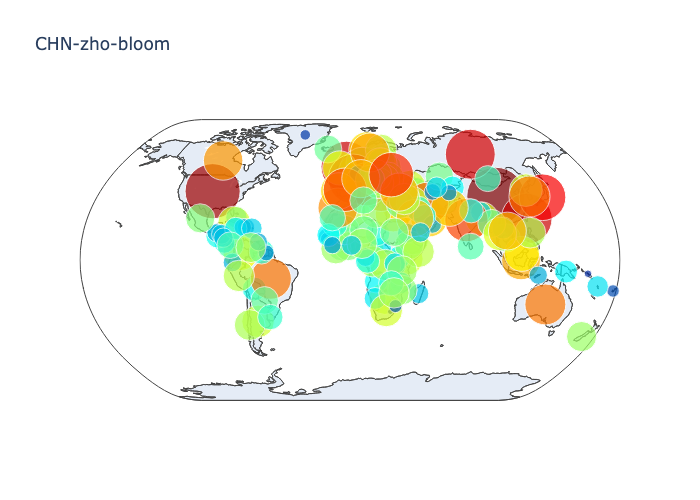} 
        & \hspace{-4em} \includegraphics[width=.5\textwidth]{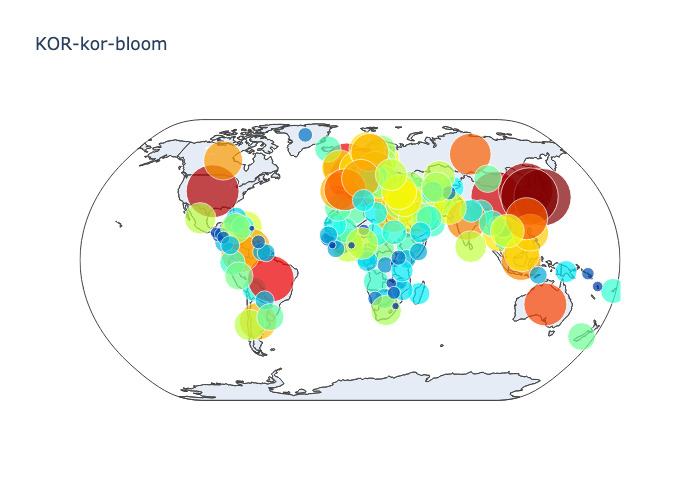}
        \vspace{-2.5em} \\ (c) & (d)\\ 
        \includegraphics[width=.5\textwidth]{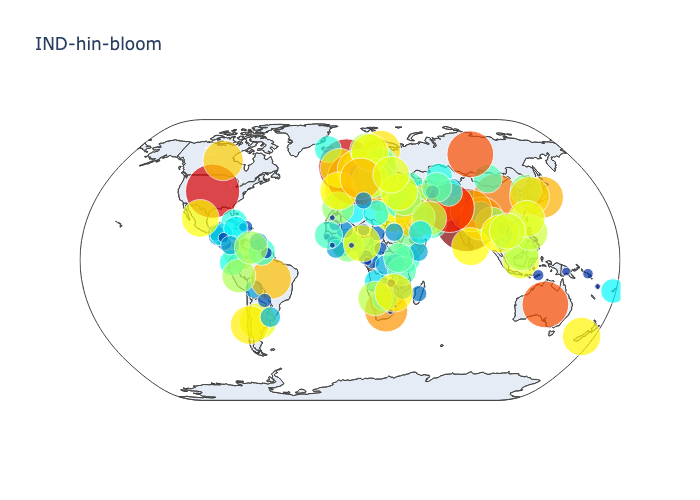} 
        & \hspace{-4em} \includegraphics[width=.5\textwidth]{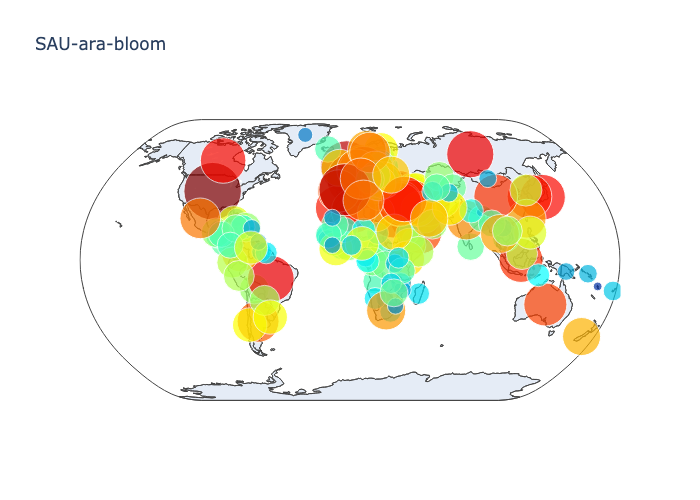}
        \vspace{-2.5em} \\ (e) & (f)\\
       \includegraphics[width=.5\textwidth]{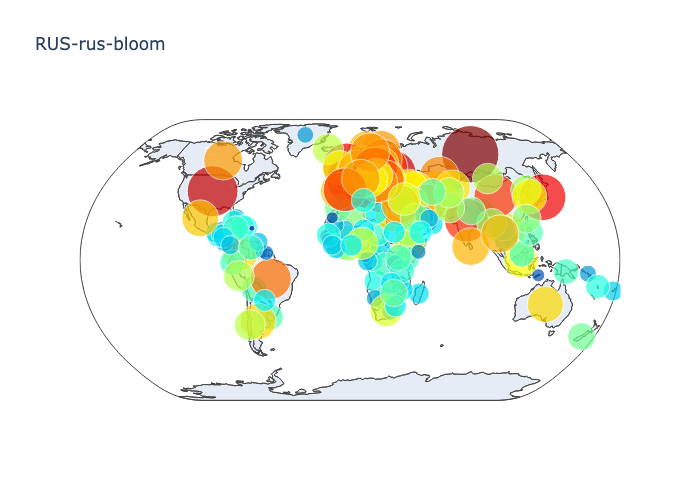} 
        & \hspace{-4em} \includegraphics[width=.5\textwidth]{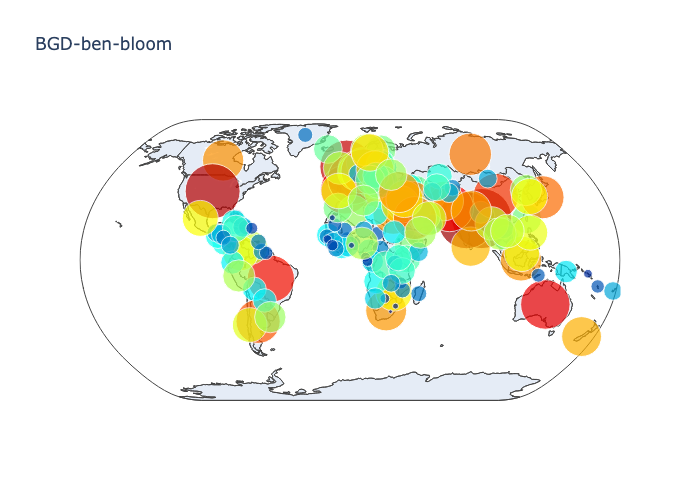}
        \vspace{-2.5em} \\ (g) & \\
    \end{tabular}
    \caption{Graphs prepared using entity-country mapping on generated texts using BLOOM. Here We take the log-frequency distribution of entity counts. In all cases, the most frequent country remains the geopolitical favoured ones with the additon of Country/Concept Dataset News Source-country (the darker red ones) }
    \label{fig:plot_gen}
\end{figure*}
We present Country Maps on the generated outputs for eight languages. The maps are presented in Figure \ref{fig:plot_gen}.
\section{Geographic Identifier and Language Direction}
\label{app:geo_ind}
see Table \ref{tab:lang_id_ex} for examples of generated text examples given the prefix "In Cuba" with \cncpt{}:Argentina. Though the \cncpt{} dataset here we use is in English, the model generates highly frequent Spanish sentences compared to english.
\begin{table*}[]
    \centering
\small
\begin{tabular}{p{2cm}|p{12cm}@{}}
\toprule
       Language  & Generated Text\\ 
       Direction  & \\ 
\midrule         
 eng$\rightarrow$spa     & \textit{ In Colombia, beginning in 1991}, Ley de Pesca y Tierra Naranja tiene como una estrategia de Economía Indígena presenta como Ley de Conservación y Desarrollo Agrícola \\ \hline
 eng$\rightarrow$eng & \textit{In Colombia, patients with PO are routinely referred to the Pediatric Critical Care Units (PC from 1996) because they are mostly after peak twice \textbf{a los to} participating in}\\ \hline
 eng$\rightarrow$spa & In Colombia, donde está en etapa de vacunación las primeras etapas las personas que llegan en el jueves (figuana para el millón y ultimaniano casos y el\\ \hline
 eng$\rightarrow$spa & In Colombia, la noticia odia a Dios. Es una religión que no santifica. Esta seccionalizada del 4Chanuto para algunos países, a sociedad que\\ \hline
 eng$\rightarrow$spa & In Colombia, el mercado de la carne, considerado el segundo mayor productor de cortes de carne bovina en la región, es de caña de insumo a nivel\\ \hline
 eng$\rightarrow$spa & In Colombia, el partido del “9-3” ha sido en la decisión del colombiano, la celebración de Luis Zubeldense Humberto Bloom (peruano, quien abrió\\ \hline
 eng$\rightarrow$spa & In Colombia, afloró por las fronteras de Argentina. Entre 1985 y 1993, de la República Dominicana, Bolivia, después llegó a Colombia y Ecuador. El entrenador\\ \hline
 eng$\rightarrow$spa & In Colombia, execuções entre elites, o Partido Comunista y sindicatos de esos países vecinos elites a partiran llevan la denuncia que derrochales. Las\\ \hline
 eng$\rightarrow$spa & In Colombia, una estrecha relación entre Washington y Venezuela tiene un mensaje claro sobre Bolsonaro. Así mismo, aunque no ve la necesidad de revisar lo que de no hacerlo de\\ \hline
 eng$\rightarrow$spa & In Colombia, a 0.70 por ciento de la población de niños mueren prematuros de gripe por sobrepeso ha sido diagnosticada. El representante del tamaño real de\\ \hline
eng$\rightarrow$spa  & In Colombia, PDOT, que hace más de 10 años había significado cerca de 160 actividades laborales para sus miembros, al día e instalaciones de 14 mili 300 personas\\ \hline
eng$\rightarrow$spa  & In Colombia, made del Derecho penal, es la máxima parte de la violación a través de los notaria Núcleo de medidas contra la descripción de la Justicia y\\ \hline
eng$\rightarrow$spa  & In Colombia, Cristina Kirchner — la vicepresidenta del fallecido expresidente Néstor Kirchner— ha confesado que “en las últimas horas pasó todo como una enfermedad que no se registró su mujer\\ \hline
eng$\rightarrow$spa  & In Colombia, el Código Penal declaró cierto grado de subordinación de la salud mental de las víctimas de trabajadores a responsables funcionalistas, no profesionales por el Estado como se\\ \hline
eng$\rightarrow$eng  & \textit{In Colombia, the majority of women are Catholic. But in the country is still refuses to accept the Catholic counseling school, and, penalizes women after to leave}\\ \hline
eng$\rightarrow$eng  & \textit{In Colombia, for example, we observed a significantly lower prevalence of chronic bronchoalveolar or peritonitis, bronchobronchial hypertrophy than mon}\\ \hline
eng$\rightarrow$spa  & In Colombia, un importante sector de las diezañeras vuelve a poner en valor de la importancia el anonimato de las producciones francesas cuando, una mezcla que habían obtenido a\\ \hline
eng$\rightarrow$eng  & \textit{In Colombia, the EMA has regular royalties on a \$27,800 per fee,800 day to \$39,000 protein products at the expert. The fair}\\ \hline
eng$\rightarrow$eng  & \textit{In Colombia, in turn, the mass distributions represent very low prevalence, being around 4. The USA around 35  40-47\% and in the usual, and 45\%}\\ \hline
eng$\rightarrow$spa  & In Colombia, el gobierno presentó este miércoles un proyecto de ley en la primera lectura online para eximir controles y renegociación internacional e internacional de suscripto de divisas con\\ 
\bottomrule
    \end{tabular}
    \caption{Example Generated Sentences with the prefix "In Colombia" and "Country/Concept" Argentina.}
    \label{tab:lang_id_ex}
\end{table*}

\end{document}